\documentclass{article} 
\usepackage{iclr2026_conference,times}

\usepackage{amsmath,amsfonts,bm}

\def\eqref#1{equation~\ref{#1}}

\def\1{\bm{1}}

\DeclareMathAlphabet{\mathsfit}{\encodingdefault}{\sfdefault}{m}{sl}
\SetMathAlphabet{\mathsfit}{bold}{\encodingdefault}{\sfdefault}{bx}{n}

\iclrfinalcopy

\makeatletter
\def\ps@headings{%
\def\@oddhead{\hfill \hfill} 
\def\@evenhead{\hfill \hfill}
\def\@oddfoot{\hfill\thepage\hfill}
\def\@evenfoot{\hfill\thepage\hfill}}
\makeatother
\usepackage{hyperref}
\usepackage{url}

\usepackage{microtype}
\usepackage{graphicx}
\usepackage{booktabs}
\usepackage{multirow}
\usepackage{amsmath, amssymb}
\usepackage{hyperref}
\usepackage{xcolor}
\usepackage{wrapfig}
\usepackage{subcaption}
\usepackage{cleveref}
\usepackage{algorithm}
\usepackage{algpseudocode}

\hypersetup{colorlinks=true, linkcolor=blue!60!black,
citecolor=blue!60!black, urlcolor=blue!60!black}

\newcommand{\method}{Vernier}

\newcommand{\Pzero}{P_0}
\newcommand{\Pone}{P_1}
\newcommand{\Lcons}{\mathcal{L}_{\mathrm{cons}}}
\newcommand{\Ltask}{\mathcal{L}_{\mathrm{task}}}

\providecommand{\stdv}[1]{{\scriptsize\textpm#1}}

\title{Vernier: Probing Representational Misalignment Behind Lexical Gaps in Causal Reasoning}

\author{Zhenyu Yu \\
College of Computer Science and Artificial Intelligence\\
Fudan University \\
\texttt{yuzhenyuyxl@foxmail.com}
}

\begin{document}

\maketitle

\begin{abstract}
Instruction-tuned language models can answer the same causal-reasoning question differently after its English variable names are replaced by type-preserving placeholders, although the structural causal model and the gold answer are unchanged. We ask whether this lexical gap reflects information loss in the placeholder view or a misaligned read-out from a representation that still carries answer-relevant content. \method{} uses a paired-view weight update as an instrument and then inspects the mechanism left after the gap closes. In the working regimes, the evidence favours representational misalignment. A variable-name probe becomes more accurate on the placeholder view, and activation patching on Qwen-7B, Qwen-14B, and Llama-3.1-8B shows that the decision-token representation can transfer answer identity between views. The update that realigns the views is counterfactual augmentation over original and placeholder prompts, while the answer-subspace KL mainly sharpens intermediate answer-belief agreement. Success is bounded by model family, scale, and task. CRASS transfer is reliable across Qwen scales and Llama, e-CARE remains weak, and preliminary non-causal rename tasks show a similar qualitative pattern.
\end{abstract}

\section{Introduction}
\label{sec:intro}

Instruction-tuned language models can give different answers to the same causal-reasoning question when its variable names change. Replacing surface forms such as \emph{smoking}, \emph{husband}, and \emph{alarm clock} with typed placeholders \(X_1, X_2, X_3\) in a structurally identical prompt produces a different answer distribution, even though the structural causal model, the joint distribution, the query, and the gold answer are all preserved. \citet{caliper} report this lexical gap across model families and find that direct prompting, chain-of-thought prompting, and scaffolded prompting leave it visible. Related 2025--2026 evaluations report similar brittleness under meaning-preserving rewrites, context shifts, and anti-semantic controls in causal or mathematical reasoning settings \citep{hao2025mathrobust,lee2025econcausal,wang2026causalflip,li2026meter}. The failure is therefore not just that the model needs a better prompt.

The scientific question is what kind of failure the lexical gap is. One possibility is an information deficit: after the names are removed, the placeholder view no longer contains the story-level content needed to answer. Another is representational misalignment: the content is still present, but the post-trained model reads it out differently from the two lexical forms. This distinction mirrors a broader line of work on spurious-signal reliance and prompt sensitivity in language models \citep{mccoy2019hans,webson2022prompts,sclar2024prompt}. These accounts matter because they imply different fixes. If the placeholder view loses information, then a successful update may simply teach the model to ignore or overwrite missing content. If the view is misaligned, a successful update should make the same item-level content more accessible from both forms.

We study this question by using a targeted weight update as an instrument. The update is not proposed as a new general-purpose fine-tuning recipe. It is a controlled way to ask what changes when the gap closes. Our method, \method{}, trains a LoRA adapter on paired lexical views \citep{lora}. Both views receive answer supervision, and an answer-subspace symmetric KL term encourages their answer distributions to agree, following the general consistency-training principle while changing the view-construction mechanism \citep{xie2020uda,rdrop}. The ablations below show that the paired lexical views, rather than generic consistency, are the main source of behavioural closure. The KL term is used to make the answer-belief alignment explicit and measurable. Recent benchmark work also motivates this design by showing that prompt-side variation alone does not reliably yield stable reasoning behavior \citep{sclar2024prompt,chi2025causalmirage}. The answer-subspace restriction matches the evaluation metric, which depends only on a small answer set rather than the full vocabulary.

The rest of the paper is organised around falsification tests for the competing explanations. If closure removes information, story-level lexical content should become less recoverable from the perturbed view. If generic consistency is enough, then dropout consistency without the lexical perturbation should help. If the two views differ by a global direction, then a map fitted once from paired hidden states should close the gap at inference. If the update merely damages capability, downstream evaluations should reveal that. We therefore combine behavioural evaluation with probes and logit-lens diagnostics \citep{belrose2023tunedlens}, activation patching \citep{zhang2024patching}, steering tests \citep{turner2023actadd,zou2023repe}, capability checks \citep{mmlu}, and model-family stress tests. This structure is intended to identify what the update repairs, beyond whether it improves a benchmark number.

\noindent\textbf{Contributions.}
\begin{itemize}
  \item \textbf{We turn a lexical robustness failure into a falsifiable mechanism question.} The paper separates an information-deficit account from a representational-misalignment account using paired-view training, probes, logit-lens diagnostics, activation patching, and inference-time steering tests. The evidence is organised so that each diagnostic can rule out a specific weaker explanation rather than merely add another benchmark number.
  \item \textbf{We introduce \method{} as a controlled realignment instrument.} \method{} trains on original and structure-preserving perturbed views with answer supervision on both and an answer-subspace consistency penalty. The paired-view augmentation is the behavioural engine, while the answer-subspace KL provides a belief-level alignment diagnostic. The method is deliberately narrow: it targets the answer set used by the causal-QA evaluation and is used to ask what changes when a lexical gap closes.
  \item \textbf{We map the boundary conditions of the effect.} The experiments cover multiple model families, a within-family Qwen scale sweep, out-of-distribution causal benchmarks, capability checks, placeholder-scheme stress tests, non-causal rename tasks, and failed global-map interventions. This lets us state where the update works and where it fails, transfers weakly, or cannot be reproduced by a single inference-time direction.
\end{itemize}

\section{Related Work}
\label{sec:related}

\subsection{Lexical robustness and causal reasoning in LLMs}

\citet{caliper} introduce the structure-preserving lexical perturbation \(T(\cdot)\) we adopt on CLadder \citep{cladder}, CRASS \citep{crass}, and e-CARE \citep{ecare}: content words that name variables are replaced by typed placeholders while the causal graph, query type, answer set, and gold answer are held fixed. The perturbation is designed to expose whether a model's causal answer depends on semantically rich surface names rather than on the structural causal query. This surface dependence is part of a broader robustness pattern. \citet{chi2025causalmirage} argue that current LLMs perform mostly shallow causal reasoning rather than genuine structural inference, \citet{hao2025mathrobust} show that renaming variables degrades mathematical reasoning by several points, and \citet{lee2025econcausal} find that causal predictions fail to track changing context. These works motivate asking not only whether a gap exists, but what internal failure it reflects. For the causal-QA case we ask whether the gap is information loss or representational misalignment, and we answer it mechanistically rather than only measuring it. The structural causal model framework of \citet{pearl2009,pearl2018} formalises the observational, interventional, and counterfactual queries the three benchmarks operationalise. Adjacent work shows that LLMs exploit surface features in ways that interact with task-relevant signal. \citet{mccoy2019hans} demonstrate syntactic-heuristic reliance on NLI. \citet{webson2022prompts} show that prompt-based zero-shot models can ignore instruction semantics while remaining sensitive to surface choices. \citet{sclar2024prompt} quantify sensitivity to prompt formatting. \citet{min2022rethinking} find that label correctness in in-context demonstrations matters less than label distribution. Chain-of-thought prompting \citep{wei2022cot,kojima2022zeroshot} and self-consistency \citep{wang2023selfconsistency} are natural prompt-side interventions. Because they do not modify the model's read-out from the two lexical views, they motivate our weight-update approach rather than replacing it.

\subsection{Invariance, consistency training, and counterfactual augmentation}

\method{} belongs to the broad family of methods that build a known invariance into a model. Where group-equivariant networks \citep{cohen2016group} hard-wire invariance to a symmetry group into the architecture, and consistency-regularised semi-supervised learning \citep{xie2020uda} softens a known data augmentation into a penalty, \method{} treats the structure-preserving perturbation \(T\) of \citet{caliper} as the task symmetry and penalises violations of it as a soft constraint on the answer distribution. The perturbation is a genuine invariance of the task by construction, which is what separates this setting from the more common one where the augmentation is only approximately label-preserving. Within consistency training, R-Drop \citep{rdrop} adds a symmetric KL between two dropout-sampled forward passes of one input and SimCSE \citep{simcse} contrasts two dropout-sampled views of one sentence, both treating the two views as stochastic copies of a single input. \method{} differs in two ways. The two views are produced by the lexical edit rather than by dropout sampling, which makes the consistency target a property of the data distribution and not of the network's internal noise, and the KL is restricted to the answer-token subspace rather than the full vocabulary. The first difference is essential because the R-Drop baseline widens the gap without \(T\). \citet{kaushik2020learning} train classifiers on human-rewritten counterfactual pairs to improve robustness to spurious correlations. Our \(\beta = 0\) ablation is the closest analogue without the KL term and closes most of the in-distribution gap but only part of the out-of-distribution gap.

\subsection{Mechanistic interpretability of hidden representations}

Our mechanism analysis uses a logit-lens-style diagnostic: intermediate hidden states are projected through the unembedding matrix and compared on the answer-token subspace. The tuned-lens variant \citep{belrose2023tunedlens} learns a small per-layer adapter for a related latent-prediction analysis. The transformer-circuits framework of \citet{elhage2021circuits} provides the conceptual vocabulary for the per-layer analysis. \citet{meng2022rome} locate factual associations to mid-layer MLP modules, and \citet{geva2023dissecting} dissect per-layer dynamics of factual recall. Our linear-probe analysis uses a single multinomial classifier per layer to test layerwise recoverability of the variable-name family from the hidden state, complementary to the causal-mediation techniques of these prior works. A related line controls model behaviour by adding a single direction to the residual stream, with activation addition \citep{turner2023actadd} and representation engineering \citep{zou2023repe} steering attributes such as sentiment, truthfulness, and refusal. We use the same construction as a test and find the causal-reasoning lexical gap does not admit such a direction, unlike those attributes, while non-causal variants are partly steerable. Instruction tuning by reinforcement learning from human feedback \citep{ouyang2022instructgpt} and by multitask prompted training \citep{sanh2022t0,wei2022flan} are the post-training stages that produce the lexical sensitivity we close. We use LoRA \citep{lora} for parameter-efficient fine-tuning, AdamW \citep{adamw} optimisation, and QLoRA \citep{qlora} for the 14B and larger models.

\begin{figure}[h]
  \centering
  \includegraphics[width=0.96\linewidth]{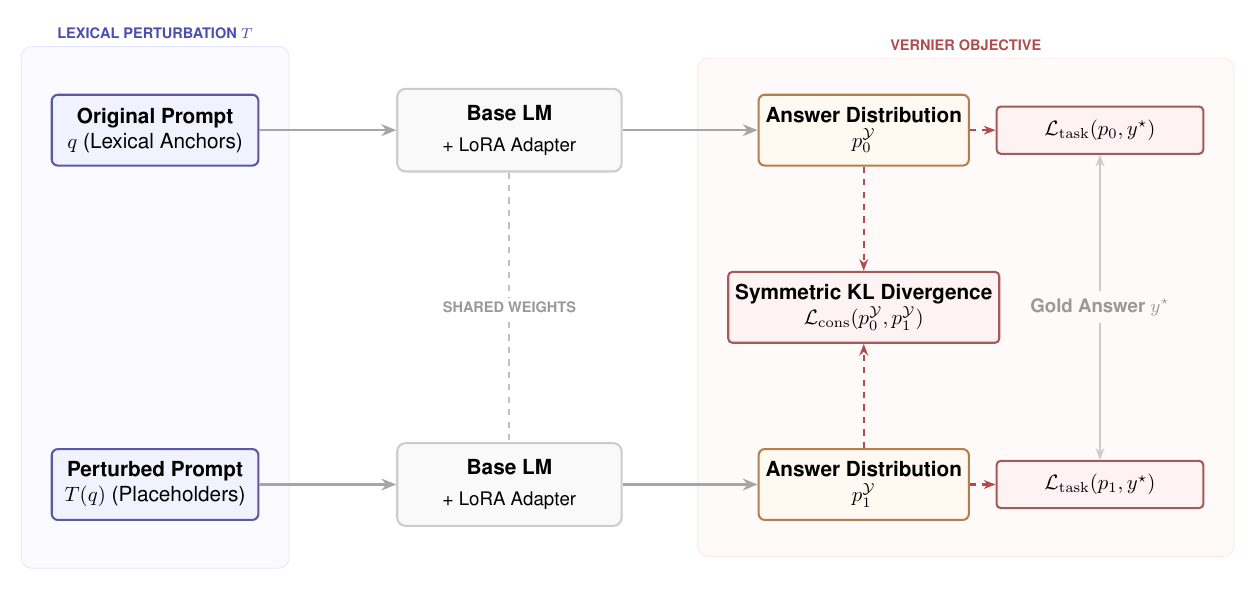}
  \caption{\method{} training schematic. The original prompt \(q\) and its structure-preserving perturbation \(T(q)\) are forwarded through the same LoRA-adapted model. The loss combines per-view answer supervision with an answer-subspace symmetric KL term.}
  \label{fig:overview}
\end{figure}

\section{Method: \method{}}
\label{sec:method}

The method has three parts (see \Cref{fig:overview}). The lexical perturbation is formalised as a structure-preserving transformation that should leave the answer distribution invariant. \method{} then trains on paired lexical views with a task loss on both views and a consistency loss on the answer-token subspace. The answer-subspace loss is in turn connected to the identifiability constraint, including the margin condition needed to translate distributional agreement into arg-max agreement. We use the update as an instrument rather than as the whole argument: the method is designed to create a controlled before/after comparison whose hidden-state consequences can then be inspected.

\subsection{Lexical perturbation and gap}
\label{sec:setup-formal}

A causal-QA item is \((q,y^\star,M)\), where \(q\) is the prompt, \(y^\star \in \mathcal{Y}\) is the gold answer, and \(M\) is the structural causal model. The lexical perturbation \(T\) replaces each SCM variable name and its variants with a typed placeholder \(X_1,X_2,\ldots\), applied consistently throughout the item. A worked instance is provided in \Cref{tab:example}. It preserves \(M\), the query, and \(y^\star\), so an SCM-grounded reasoner should satisfy \(R(\cdot\mid q)=R(\cdot\mid T(q))\). We score a model by the signed lexical gap
\begin{equation}
  \Delta(\theta;\mathcal{D}) =
  \mathbb{E}_{(q,y^\star)\sim\mathcal{D}}
  [\mathbf{1}\{\hat y_\theta(q)=y^\star\}-
   \mathbf{1}\{\hat y_\theta(T(q))=y^\star\}],
  \label{eq:gap}
\end{equation}
where predictions use the answer-token distribution at the first generation position.

The perturbation is deliberately conservative. It does not paraphrase the story, change numbers, alter the causal graph, or change the requested answer type. It only removes the English variable names that can act as lexical anchors. This matters for interpretation: a remaining gap cannot be attributed to a changed SCM or a changed label, while a closed gap cannot by itself prove deeper causal understanding. The method therefore asks a narrower question: given two views that should be equivalent for the task, can a small weight update make the model read out the same answer-relevant information from both?

\begin{table}[t]
\centering
\caption{A worked CLadder example. Gold answer \(y^\star = \text{no}\), Pearl rung observational. Underlined surface forms in \(q\) are mapped by \(T\) to placeholders. Numeric quantities, query, and answer are unchanged.}
\label{tab:example}
\small
\begin{tabular}{@{}p{0.46\linewidth}@{\hspace{0.04\linewidth}}p{0.46\linewidth}@{}}
\toprule
\textbf{Original prompt \(q\)} & \textbf{Perturbed prompt \(T(q)\)} \\
\midrule
The overall probability of alarm set by \underline{husband} is 77\%. For \underline{husbands} that don't set the \underline{alarm}, the probability of ringing \underline{alarm} is 26\%. For \underline{husbands} that set the \underline{alarm}, the probability of ringing \underline{alarm} is 76\%. Is ringing \underline{alarm} less likely than silent \underline{alarm} overall? Answer (yes or no):
&
The overall probability of \(X_3\) set by \(X_2\) is 77\%. For \(X_2\) that don't set the \(X_3\), the probability of ringing \(X_3\) is 26\%. For \(X_2\) that set the \(X_3\), the probability of ringing \(X_3\) is 76\%. Is ringing \(X_3\) less likely than silent \(X_3\) overall? Answer (yes or no):
\\
\bottomrule
\end{tabular}
\end{table}

\subsection{\method{} objective}
\label{sec:loss}

\method{} trains a LoRA adapter on paired views \((\Pzero,\Pone)=(q,T(q))\). The original view \(\Pzero\) keeps the lexical anchors seen by the base model, while the perturbed view \(\Pone\) removes them while preserving the structural causal problem. Training on both views prevents the update from solving the task by specialising only to one surface form. The task loss makes each view answerable on its own, while the consistency term penalises disagreement between the two answer distributions:
\begin{equation}
  \mathcal{L} =
  \alpha[\Ltask(\Pzero)+\Ltask(\Pone)] +
  \beta\,\Lcons(\Pzero,\Pone),
  \quad
  \Lcons =
  \tfrac{1}{2}\mathrm{KL}(p_{\Pzero}^{\mathcal{Y}}\Vert p_{\Pone}^{\mathcal{Y}})
  +\tfrac{1}{2}\mathrm{KL}(p_{\Pone}^{\mathcal{Y}}\Vert p_{\Pzero}^{\mathcal{Y}}).
  \label{eq:vernier-loss}
\end{equation}
Here \(p_{\Pzero}^{\mathcal{Y}}\) and \(p_{\Pone}^{\mathcal{Y}}\) are the softmax distributions after restricting the next-token logits to the finite answer set \(\mathcal{Y}\). For binary yes/no CLadder items, \(\mathcal{Y}\) contains the answer tokens for \emph{yes} and \emph{no}, and for multiple-choice CRASS and e-CARE items it contains the option tokens. We use \(\alpha=\beta=1\). The loss is applied at the first generation position, and only LoRA parameters are updated. One training step is given in \Cref{alg:vernier}.

The answer-subspace restriction is a design choice, not a computational shortcut. A full-vocabulary KL would require two prompts with different surface forms to agree on every next-token continuation, including continuations that restate the variable names. That target is stronger than the task invariance and can penalise harmless lexical differences. By contrast, the answer-subspace KL matches the evaluation interface: if two views encode the same causal problem, the distribution over admissible answers should agree even if the unrestricted language-model distribution does not.

\begin{algorithm}[t]
\caption{\method{} training (one outer step).}
\label{alg:vernier}
\begin{algorithmic}[1]
\Require base LM \(f_\theta\) with frozen weights, LoRA adapter \(\phi\), answer-token id set \(\mathcal{Y}\), batch \(\{(q_i,T(q_i),y_i^\star)\}_{i=1}^B\).
\State \(\Pzero,\Pone \gets\) tokenised original and perturbed prompts.
\State \(\mathbf{z}^{(v)} \gets f_{\theta+\phi}(P_v)\) for \(v\in\{0,1\}\) at the first generation position.
\State \(\mathbf{p}^{(v)} \gets \mathrm{softmax}(\mathbf{z}^{(v)}_{\cdot,\mathcal{Y}})\).
\State \(\mathcal{L}_{\mathrm{task}} \gets -B^{-1}\sum_i[\log \mathbf{p}^{(0)}_{i,y_i^\star}+\log \mathbf{p}^{(1)}_{i,y_i^\star}]\).
\State \(\mathcal{L}_{\mathrm{cons}} \gets (2B)^{-1}\sum_i[\mathrm{KL}(\mathbf{p}^{(0)}_i\Vert\mathbf{p}^{(1)}_i)+\mathrm{KL}(\mathbf{p}^{(1)}_i\Vert\mathbf{p}^{(0)}_i)]\).
\State Update only \(\phi\) with AdamW on \(\alpha\mathcal{L}_{\mathrm{task}}+\beta\mathcal{L}_{\mathrm{cons}}\).
\end{algorithmic}
\end{algorithm}

\subsection{Distinguishing misalignment from information loss}
\label{sec:instrument}

\method{} is useful because the two explanations introduced above predict different after-effects. Under a simple information-deficit account, the perturbed prompt lacks answer-relevant story content. A successful update could still close the gap, but it would have to do so by adding a shortcut, memorising an answer pattern, suppressing lexical information, or regressing the original view. Under a representational-misalignment account, the perturbed view still carries item-level content, but the post-trained read-out fails to use it in the same way as the original view. A successful update should then make the two views more similar at the decision interface while preserving, or increasing, recoverability of the story-level lexical family from the perturbed representation.

This is why the method is paired with mechanism tests rather than evaluated only by accuracy. The primary behavioural signature is not merely a smaller gap, but a smaller gap with \(\Pone\) improving and \(\Pzero\) not collapsing. The mechanism signature is stronger: cosine and logit-lens agreement should improve at the decision token, a probe trained on \(\Pzero\) should transfer better to \(\Pone\), and activation patching should show that the decision-token representation can carry answer identity across views. These are not assumptions of the method. They are the falsification tests that follow from using the method as an instrument.

\subsection{Answer-subspace consistency}
\label{sec:identif}

The answer-subspace KL is an empirical plugin estimate of the identifiability violation on \(\mathcal{Y}\). Let \(D_{\mathrm{SKL}}\) be the symmetric KL and let
\begin{equation}
G(\theta;\mathcal{D})=\mathbb{E}_{q\sim\mathcal{D}}
[D_{\mathrm{SKL}}(p_\theta(\cdot\mid q)_{\mathcal{Y}}
\Vert p_\theta(\cdot\mid T(q))_{\mathcal{Y}})].
\label{eq:answer-subspace-kl}
\end{equation}
The batch mean of \(\Lcons\) is an unbiased estimate of \(G\). Since symmetric KL is positive definite on the finite simplex, \(G=0\) iff the two answer-token distributions agree almost everywhere. Distributional agreement then implies equal arg-max predictions, but away from zero this only controls prediction flips under a margin condition. If \(p_\theta(\cdot\mid q)_{\mathcal{Y}}\) has top-two margin at least \(\gamma\), then a flip requires total variation at least \(\gamma/2\), and Pinsker plus Jensen gives
\begin{equation}
  \Pr[\hat y_\theta(q)\neq \hat y_\theta(T(q))]
  \le \frac{\sqrt{2G(\theta;\mathcal{D})}}{\gamma}.
\label{eq:margin-flip-bound}
\end{equation}
We do not treat this bound as the load-bearing argument because near-tied answer distributions can violate the margin condition. Instead, the load-bearing evidence is behavioural closure together with the mechanism tests, which distinguish genuine \(\Pone\) improvement from degenerate closure through \(\Pzero\) regression (see \Cref{sec:mech}). The task loss prevents the trivial constant-output solution that would satisfy the consistency term with zero useful accuracy.

\subsection{Identification limits}
\label{sec:method-limits}

The method does not prove that the model has learned a human-interpretable causal algorithm. It only tests whether the lexical gap is removable by a small paired-view update and what kind of internal change accompanies that removal. Several outcomes would falsify the misalignment reading, including closure driven by \(\Pzero\) regression, lower probe transfer from \(\Pone\), no movement in decision-token agreement, or a generic consistency baseline such as R-Drop producing the same effect without ever seeing \(T(q)\). These tests are summarised in \Cref{tab:falsifiers}. We observe some of these failure signatures on Phi-3 and Gemma, which is why the later claims are scoped by model family and capacity. Conversely, when \method{} closes the gap with \(\Pone\) improvement and the mechanism diagnostics move in the predicted direction, the result supports a narrower conclusion. In that regime, the answer-relevant content was not simply erased by replacing names with placeholders.

\section{Experiments}
\label{sec:exp}

\subsection{Experimental setup}
\label{sec:setup}

\paragraph{Datasets.}

CLadder \citep{cladder} is our in-distribution benchmark. We use items 0 through 1499 of the commonsensical split for training and a disjoint slice of items 1500 through 1699 (\(n = 200\)) as the held-out evaluation set. CRASS \citep{crass} and e-CARE \citep{ecare} are out-of-distribution. No item from either is seen during training. We use 200-item samples from each. For general-capability evaluation we sample 200 items each from MMLU \citep{mmlu}, HellaSwag, and GSM8K under a fixed seed. CLadder also contains a nonsensical subset with fictional-concept variable names, used for the falsifiability test. To probe whether the account may extend beyond causal reasoning, we add two non-causal BBH tasks \citep{bbh}, logical\_deduction and reasoning\_about\_colored\_objects, with object names renamed to typed placeholders.

\paragraph{Baselines.}
We compare \method{} (\(\beta = 1\), answer-subspace KL) against five baselines on Qwen2.5-7B at step 500.
\textbf{Base.} The pretrained instruction-tuned model with no fine-tuning.
\textbf{\(\bm{\Pzero}\)-only SFT.} Supervised fine-tuning on the original prompt alone, with no perturbed view.
\textbf{R-Drop} \citep{rdrop}. Symmetric KL between two dropout-sampled forward passes of \(\Pzero\), with no perturbation \(T\).
\textbf{Twin task} (\(\beta = 0\)). Both views routed through the task loss, with no consistency term.
\textbf{Full-vocabulary KL} (\(\beta = 1\)). The same training as \method{} but with the KL computed over the full vocabulary rather than the answer subspace.

\paragraph{Metrics.}

For each item we score \((q, T(q))\) by taking the softmax over the answer-token subset at the first generation position. The primary metric is the signed lexical gap \(\Pzero - \Pone\) in percentage points, reported alongside \(\Pzero\) and \(\Pone\) accuracy so that closure driven by \(\Pone\) improvement can be distinguished from closure driven by \(\Pzero\) regression. A gap at or near zero is the goal. A negative gap is an inversion in which \(\Pone\) exceeds \(\Pzero\), which we report as such rather than as a larger success. For OOD benchmarks we also give the relative gap reduction \(\Delta_{{gap}} = ({gap}_{{Vernier}} - {gap}_{{base}}) / {gap}_{{base}}\) as a convenience, but we caution that this ratio is unstable when the base gap is small and can exceed 100 percent when the gap crosses zero, so we anchor all claims on the signed gap in percentage points rather than on this ratio. On general-capability benchmarks we report top-1 accuracy.

\paragraph{Implementation.}

We attach a LoRA \citep{lora} adapter of rank \(r = 16\) to the query, key, value, and output projections of every attention layer, with LoRA alpha 32 and dropout 0.05. We train with AdamW \citep{adamw} at learning rate \(1 \times 10^{-4}\), weight decay 0, batch size 4 pairs per micro-step, gradient accumulation 4, linear warmup over 50 steps, cosine decay thereafter, and a total budget of 500 steps. We evaluate nine instruction-tuned models spanning five families, namely Phi-3-mini-4k-instruct (3.8B) \citep{phi3}, Qwen2.5 at 1.5B, 3B, 7B, 14B, and 32B \citep{qwen25}, Llama-3.1-8B-Instruct \citep{llama3}, Mistral-7B-Instruct-v0.3 \citep{jiang2023mistral}, and Gemma-2-2B-it \citep{gemma2}. The 1.5B to 8B models train in BF16 on a single 24 GB GPU. The 14B and 32B models use QLoRA \citep{qlora} on a 40 GB GPU, the 32B with batch size 2 and gradient accumulation 8. Phi-3 trains at learning rate \(5 \times 10^{-6}\) with early stopping at step 70, since the canonical learning rate drives its adapter to a degenerate constant-output minimum by step 50. On Mistral-7B the canonical learning rate produces an aggressive in-distribution closure with substantial downstream-capability degradation. We rescue it with the Phi-3 protocol (lr=$5\!\times\!10^{-6}$, early stopping at step 150) and report both rows in the main tables. The hyperparameter trade-off is analysed with the failure modes. The full recipe-and-selection audit gives the metric used to pick each rescued checkpoint (see \Cref{sec:audit}).

\subsection{Comparison}
\label{sec:main}

\begin{table}[t]
\centering
\providecommand{\stdv}[1]{{\footnotesize\textpm#1}}
\caption{In-distribution (CLadder) and out-of-distribution (CRASS, e-CARE) held-out results, $n = 200$ each, grouped by dataset. Adapters are trained only on CLadder. The gap is $P_0 - P_1$ in percentage points and $\Delta_{\text{gap}}$ is the relative gap reduction versus base (OOD only; in-distribution base gaps are too small for a stable ratio). All $\pm$ are $1$ s.d.: for the Vernier rows of Qwen-7B, Qwen-14B, and Llama-3.1-8B ($\dagger$) this is the std across three rounds; for every other cell (all base rows and the single-seed Vernier rows of Phi-3, Qwen-32B, and Mistral-7B) it is the paired bootstrap standard error over the $200$ items. Within each dataset block, \textbf{bold} marks the best and \underline{underline} the second best per column ($P_0$/$P_1$ higher is better; Gap closest to $0$ is better; $\Delta_{\text{gap}}$ more negative is better). Qwen-32B uses QLoRA; Mistral-7B ($\ddagger$) is reported under the default and rescued recipes.}
\label{tab:main}
\small
\setlength{\tabcolsep}{4pt}
\resizebox{\textwidth}{!}{%
\begin{tabular}{c l ccc ccc c}
\toprule
& & \multicolumn{3}{c}{\textbf{Base}} & \multicolumn{3}{c}{\textbf{Vernier}} & \\
\cmidrule(lr){3-5} \cmidrule(lr){6-8}
\textbf{Data} & \textbf{Model} & $\bm{P_0}$ & $\bm{P_1}$ & \textbf{Gap} & $\bm{P_0}$ & $\bm{P_1}$ & \textbf{Gap} & \textbf{Rel. gap} \\
\midrule
\multirow{7}{*}{\rotatebox{90}{CLadder (ID)}}
  & Phi-3-mini (3.8B)              & 0.585\stdv{0.035} & 0.540\stdv{0.035} & +4.5\stdv{2.3} & 0.550\stdv{0.035} & 0.550\stdv{0.035} & \textbf{0.0\stdv{2.1}} & --- \\
  & Qwen2.5-7B$^\dagger$           & \textbf{0.625\stdv{0.034}} & 0.545\stdv{0.035} & +8.0\stdv{2.6} & 0.868\stdv{0.003} & 0.893\stdv{0.010} & -2.5\stdv{0.9} & --- \\
  & Qwen2.5-14B$^\dagger$          & \underline{0.615\stdv{0.035}} & \underline{0.580\stdv{0.035}} & +3.5\stdv{2.7} & 0.955\stdv{0.029} & 0.950\stdv{0.013} & +0.5\stdv{1.4} & --- \\
  & Qwen2.5-32B (QLoRA)            & 0.565\stdv{0.035} & \textbf{0.585\stdv{0.035}} & \textbf{-2.0\stdv{2.1}} & \underline{0.970\stdv{0.012}} & \underline{0.965\stdv{0.013}} & 0.5\stdv{1.3} & --- \\
  & Llama-3.1-8B$^\dagger$         & 0.540\stdv{0.035} & 0.515\stdv{0.035} & \underline{+2.5\stdv{3.6}} & 0.945\stdv{0.005} & 0.948\stdv{0.008} & \underline{-0.3\stdv{1.0}} & --- \\
  & Mistral-7B$^\ddagger$ (lr=1e-4)        & 0.610\stdv{0.034} & 0.565\stdv{0.035} & +4.5\stdv{3.9} & \textbf{0.980\stdv{0.010}} & \textbf{0.995\stdv{0.005}} & -1.5\stdv{1.1} & --- \\
  & Mistral-7B$^\ddagger$ (lr=5e-6, s150)  & 0.610\stdv{0.034} & 0.565\stdv{0.035} & +4.5\stdv{3.9} & 0.605\stdv{0.035} & 0.600\stdv{0.035} & +0.5\stdv{3.1} & --- \\
\midrule
\multirow{7}{*}{\rotatebox{90}{CRASS (OOD)}}
  & Phi-3-mini (3.8B)              & 0.845\stdv{0.026} & 0.545\stdv{0.035} & +30.0\stdv{3.9} & 0.840\stdv{0.026} & 0.550\stdv{0.035} & +29.0\stdv{3.9} & -3\% \\
  & Qwen2.5-7B$^\dagger$           & 0.845\stdv{0.026} & 0.505\stdv{0.035} & +34.0\stdv{3.9} & 0.870\stdv{0.013} & 0.622\stdv{0.012} & +24.8\stdv{0.3} & \underline{-27\%} \\
  & Qwen2.5-14B$^\dagger$          & \underline{0.895\stdv{0.022}} & \underline{0.625\stdv{0.034}} & \textbf{+27.0\stdv{3.5}} & \underline{0.918\stdv{0.008}} & \underline{0.722\stdv{0.013}} & \textbf{+19.7\stdv{0.8}} & \underline{-27\%} \\
  & Qwen2.5-32B (QLoRA)            & \textbf{0.910\stdv{0.020}} & \textbf{0.635\stdv{0.034}} & \underline{+27.5\stdv{3.6}} & \textbf{0.950\stdv{0.015}} & \textbf{0.725\stdv{0.031}} & +22.5\stdv{3.3} & -18\% \\
  & Llama-3.1-8B$^\dagger$         & 0.850\stdv{0.025} & 0.445\stdv{0.035} & +40.5\stdv{3.8} & 0.872\stdv{0.008} & 0.558\stdv{0.035} & +31.3\stdv{3.1} & -23\% \\
  & Mistral-7B$^\ddagger$ (lr=1e-4)        & 0.795\stdv{0.028} & 0.415\stdv{0.035} & +38.0\stdv{3.9} & 0.710\stdv{0.032} & 0.500\stdv{0.036} & \underline{+21.0\stdv{3.4}} & \textbf{-45\%} \\
  & Mistral-7B$^\ddagger$ (lr=5e-6, s150)  & 0.795\stdv{0.028} & 0.415\stdv{0.035} & +38.0\stdv{3.9} & 0.735\stdv{0.031} & 0.365\stdv{0.034} & +37.0\stdv{3.8} & -3\% \\
\midrule
\multirow{7}{*}{\rotatebox{90}{e-CARE (OOD)}}
  & Phi-3-mini (3.8B)              & 0.795\stdv{0.029} & 0.590\stdv{0.035} & \underline{+20.5\stdv{3.9}} & 0.785\stdv{0.029} & 0.600\stdv{0.035} & \underline{+18.5\stdv{3.9}} & -10\% \\
  & Qwen2.5-7B$^\dagger$           & \underline{0.815\stdv{0.028}} & 0.585\stdv{0.035} & +23.0\stdv{3.8} & 0.810\stdv{0.010} & \textbf{0.622\stdv{0.012}} & +18.8\stdv{1.5} & \underline{-18\%} \\
  & Qwen2.5-14B$^\dagger$          & \textbf{0.820\stdv{0.027}} & \underline{0.605\stdv{0.035}} & +21.5\stdv{3.7} & \textbf{0.823\stdv{0.002}} & 0.610\stdv{0.004} & +21.3\stdv{0.5} & -1\% \\
  & Qwen2.5-32B (QLoRA)            & \textbf{0.820\stdv{0.027}} & \textbf{0.625\stdv{0.034}} & \textbf{+19.5\stdv{3.8}} & \underline{0.815\stdv{0.028}} & 0.595\stdv{0.035} & +22.0\stdv{3.9} & +13\% \\
  & Llama-3.1-8B$^\dagger$         & 0.790\stdv{0.029} & 0.550\stdv{0.035} & +24.0\stdv{4.5} & 0.793\stdv{0.005} & \underline{0.615\stdv{0.004}} & \textbf{+17.8\stdv{0.6}} & \textbf{-26\%} \\
  & Mistral-7B$^\ddagger$ (lr=1e-4)        & 0.790\stdv{0.029} & 0.555\stdv{0.035} & +23.5\stdv{4.1} & 0.775\stdv{0.030} & 0.580\stdv{0.035} & +19.5\stdv{4.2} & -17\% \\
  & Mistral-7B$^\ddagger$ (lr=5e-6, s150)  & 0.790\stdv{0.029} & 0.555\stdv{0.035} & +23.5\stdv{4.1} & 0.770\stdv{0.030} & 0.565\stdv{0.035} & +20.5\stdv{3.6} & -13\% \\
\bottomrule
\end{tabular}
}
\end{table}

\begin{table}[t]
\centering
\providecommand{\stdv}[1]{{\footnotesize\textpm#1}}
\caption{Out-of-distribution results on CRASS and e-CARE ($n = 200$ each), grouped by dataset, with adapters trained only on CLadder. $\Delta_{\text{gap}}$ is the relative gap reduction versus base. All $\pm$ are $1$ s.d.: for the Vernier rows of Qwen-7B, Qwen-14B, and Llama-3.1-8B ($\dagger$) it is the std over three rounds; for every other cell (all base rows and the single-seed Vernier rows of Phi-3, Qwen-32B, and Mistral-7B) it is the paired bootstrap standard error over the $200$ items. Within each dataset block, \textbf{bold} marks the best and \underline{underline} the second best per column ($P_0$/$P_1$ higher is better; Gap closest to $0$ is better; $\Delta_{\text{gap}}$ more negative is better). Qwen-32B uses QLoRA; Mistral-7B ($\ddagger$) is reported under the default and rescued recipes.}
\label{tab:ood}
\small
\setlength{\tabcolsep}{4pt}
\resizebox{\textwidth}{!}{%
\begin{tabular}{c l ccc ccc c}
\toprule
& & \multicolumn{3}{c}{\textbf{Base}} & \multicolumn{3}{c}{\textbf{Vernier}} & \\
\cmidrule(lr){3-5} \cmidrule(lr){6-8}
\textbf{Data} & \textbf{Model} & $\bm{P_0}$ & $\bm{P_1}$ & \textbf{Gap} & $\bm{P_0}$ & $\bm{P_1}$ & \textbf{Gap} & \textbf{Rel. gap} \\
\midrule
\multirow{7}{*}{\rotatebox{90}{CRASS}}
  & Phi-3-mini (3.8B)              & 0.845\stdv{0.026} & 0.545\stdv{0.035} & +30.0\stdv{3.9} & 0.840\stdv{0.026} & 0.550\stdv{0.035} & +29.0\stdv{3.9} & -3\% \\
  & Qwen2.5-7B$^\dagger$           & 0.845\stdv{0.026} & 0.505\stdv{0.035} & +34.0\stdv{3.9} & 0.870\stdv{0.013} & 0.622\stdv{0.012} & +24.8\stdv{0.3} & \underline{-27\%} \\
  & Qwen2.5-14B$^\dagger$          & \underline{0.895\stdv{0.022}} & \underline{0.625\stdv{0.034}} & \textbf{+27.0\stdv{3.5}} & \underline{0.918\stdv{0.008}} & \underline{0.722\stdv{0.013}} & \textbf{+19.7\stdv{0.8}} & \underline{-27\%} \\
  & Qwen2.5-32B (QLoRA)            & \textbf{0.910\stdv{0.020}} & \textbf{0.635\stdv{0.034}} & \underline{+27.5\stdv{3.6}} & \textbf{0.950\stdv{0.015}} & \textbf{0.725\stdv{0.031}} & +22.5\stdv{3.3} & -18\% \\
  & Llama-3.1-8B$^\dagger$         & 0.850\stdv{0.025} & 0.445\stdv{0.035} & +40.5\stdv{3.8} & 0.872\stdv{0.008} & 0.558\stdv{0.035} & +31.3\stdv{3.1} & -23\% \\
  & Mistral-7B$^\ddagger$ (lr=1e-4)        & 0.795\stdv{0.028} & 0.415\stdv{0.035} & +38.0\stdv{3.9} & 0.710\stdv{0.032} & 0.500\stdv{0.036} & \underline{+21.0\stdv{3.4}} & \textbf{-45\%} \\
  & Mistral-7B$^\ddagger$ (lr=5e-6, s150)  & 0.795\stdv{0.028} & 0.415\stdv{0.035} & +38.0\stdv{3.9} & 0.735\stdv{0.031} & 0.365\stdv{0.034} & +37.0\stdv{3.8} & -3\% \\
\midrule
\multirow{7}{*}{\rotatebox{90}{e-CARE}}
  & Phi-3-mini (3.8B)              & 0.795\stdv{0.029} & 0.590\stdv{0.035} & \underline{+20.5\stdv{3.9}} & 0.785\stdv{0.029} & 0.600\stdv{0.035} & \underline{+18.5\stdv{3.9}} & -10\% \\
  & Qwen2.5-7B$^\dagger$           & \underline{0.815\stdv{0.028}} & 0.585\stdv{0.035} & +23.0\stdv{3.8} & 0.810\stdv{0.010} & \textbf{0.622\stdv{0.012}} & +18.8\stdv{1.5} & \underline{-18\%} \\
  & Qwen2.5-14B$^\dagger$          & \textbf{0.820\stdv{0.027}} & \underline{0.605\stdv{0.035}} & +21.5\stdv{3.7} & \textbf{0.823\stdv{0.002}} & 0.610\stdv{0.004} & +21.3\stdv{0.5} & -1\% \\
  & Qwen2.5-32B (QLoRA)            & \textbf{0.820\stdv{0.027}} & \textbf{0.625\stdv{0.034}} & \textbf{+19.5\stdv{3.8}} & \underline{0.815\stdv{0.028}} & 0.595\stdv{0.035} & +22.0\stdv{3.9} & +13\% \\
  & Llama-3.1-8B$^\dagger$         & 0.790\stdv{0.029} & 0.550\stdv{0.035} & +24.0\stdv{4.5} & 0.793\stdv{0.005} & \underline{0.615\stdv{0.004}} & \textbf{+17.8\stdv{0.6}} & \textbf{-26\%} \\
  & Mistral-7B$^\ddagger$ (lr=1e-4)        & 0.790\stdv{0.029} & 0.555\stdv{0.035} & +23.5\stdv{4.1} & 0.775\stdv{0.030} & 0.580\stdv{0.035} & +19.5\stdv{4.2} & -17\% \\
  & Mistral-7B$^\ddagger$ (lr=5e-6, s150)  & 0.790\stdv{0.029} & 0.555\stdv{0.035} & +23.5\stdv{4.1} & 0.770\stdv{0.030} & 0.565\stdv{0.035} & +20.5\stdv{3.6} & -13\% \\
\bottomrule
\end{tabular}
}
\end{table}

In-distribution gap closure is reported in \Cref{tab:main}. \method{} drives the gap to at or near zero on all six models in the table. On Qwen-7B, Qwen-14B, and Llama-3.1-8B both \(\Pzero\) and \(\Pone\) rise by 24 to 44 pp from base, so the closure is driven by joint accuracy gains rather than by a regression on \(\Pzero\). Mistral-7B shows the same joint gains under the default learning rate but at a steep capability cost, so we report its rescued low-rate row as the operating point. On Phi-3 the closure has the opposite signature, with \(\Pzero\) falling by 3.5 pp and \(\Pone\) rising by only 1.0 pp. This model-specific failure is analysed with the mechanism and scale diagnostics. The Qwen2.5-32B base model already has a negative gap of \(-2.0\) pp on this slice, so the gap-closure metric is not directly comparable, but both \(\Pzero\) and \(\Pone\) rise by roughly 40 pp after \method. The in-distribution base gaps are small on larger held-out CLadder slices, and their paired confidence intervals cross zero for the base models (see \Cref{tab:base-gap-significance}). We therefore treat in-distribution closure as a sanity check and rest the behavioural case on the out-of-distribution transfer below.

Zero-shot transfer to CRASS and e-CARE is reported in \Cref{tab:ood}. The adapter trained only on CLadder reduces the CRASS gap by 18 to 27 percent across the three Qwen scales, with the Qwen-7B value averaged over three seeds. The larger-OOD check strengthens this CRASS result rather than weakening it (see \Cref{tab:ood-bign}). On all 274 prepared CRASS items, the gap falls from 35.0 to 23.7 pp on Qwen-7B and from 27.4 to 16.8 pp on Qwen-14B. The e-CARE picture on Qwen is weaker and non-monotonic in scale. At \(n=200\), Qwen-7B reduces e-CARE by 18 percent, Qwen-14B is within sampling noise (\(-2\) percent), and Qwen-32B widens the e-CARE gap by 2.5 pp despite closing CLadder and CRASS. At \(n=1000\), the e-CARE reduction remains small, from 20.7 to 18.7 pp on Qwen-7B and from 23.2 to 22.4 pp on Qwen-14B (see \Cref{tab:ood-bign}). Llama-3.1-8B reduces the CRASS gap by 23 percent and the e-CARE gap by 26 percent as a three-seed mean. The rescued Mistral-7B (lr=$5\!\times\!10^{-6}$, step 150) shows a 13 percent e-CARE reduction but only 3 percent on CRASS. Phi-3 transfer falls within noise on both benchmarks, the same null we see on its mechanism diagnostics.

\subsection{Ablation study}
\label{sec:ablation}

The ingredients are separated in \Cref{tab:ablation} on Qwen-7B using five single-seed reference configurations at step 500. The headline three-seed numbers remain in \Cref{tab:main,tab:ood}. \(\Pzero\)-only SFT widens the in-distribution gap from 8.0 to 9.5 pp. Training only on the original view reinforces the lexical anchors. R-Drop widens the gap further to 18.5 pp, with \(\Pzero\) rising to 0.965 and \(\Pone\) staying at 0.780. Dropout consistency makes the model more confident under its own noise, but it never trains on the perturbed view. The perturbation \(T\), not generic consistency, is what matters. The twin-view task row (\(\beta = 0\)) is counterfactual data augmentation in the sense of \citet{kaushik2020learning}. Both views enter the task loss, with no KL term. This row drops the in-distribution gap to 1.0 pp and gives 21 percent CRASS and 13 percent e-CARE reductions. The answer-subspace KL adds little behaviourally. On the larger CRASS evaluation (\(n = 274\)), augmentation alone reduces the Qwen-7B gap from 34.3 to 18.2 pp, while the full \method{} loss reaches 23.4 pp. The same larger-CRASS comparison on Qwen-14B gives nearly identical reductions for twin task and \method{} (see \Cref{tab:crass-aug-kl}). Full-vocabulary KL trades the other way, improving CRASS but nearly losing e-CARE. The KL term's clearest contribution is mechanistic. It sharpens logit-lens agreement between the two views beyond augmentation alone, without changing accuracy, probe transfer, or cosine geometry. We treat it as a belief-level refinement, not as the source of closure.

\subsection{Mechanism analysis}
\label{sec:mech}

\begin{figure}[t]
  \centering
  \begin{minipage}[t]{0.24\linewidth}
    \centering
    \includegraphics[width=\linewidth]{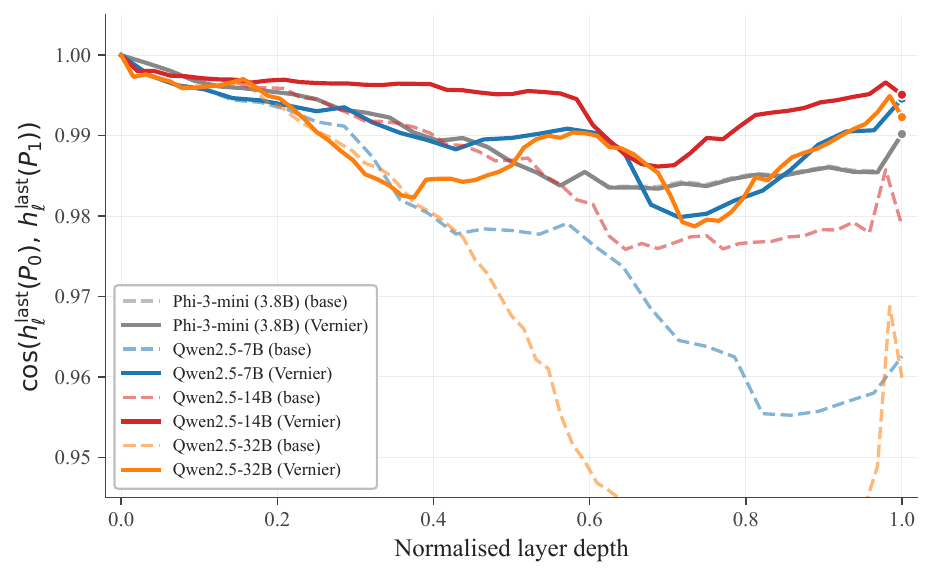}
  \end{minipage}\hfill
  \begin{minipage}[t]{0.24\linewidth}
    \centering
    \includegraphics[width=\linewidth]{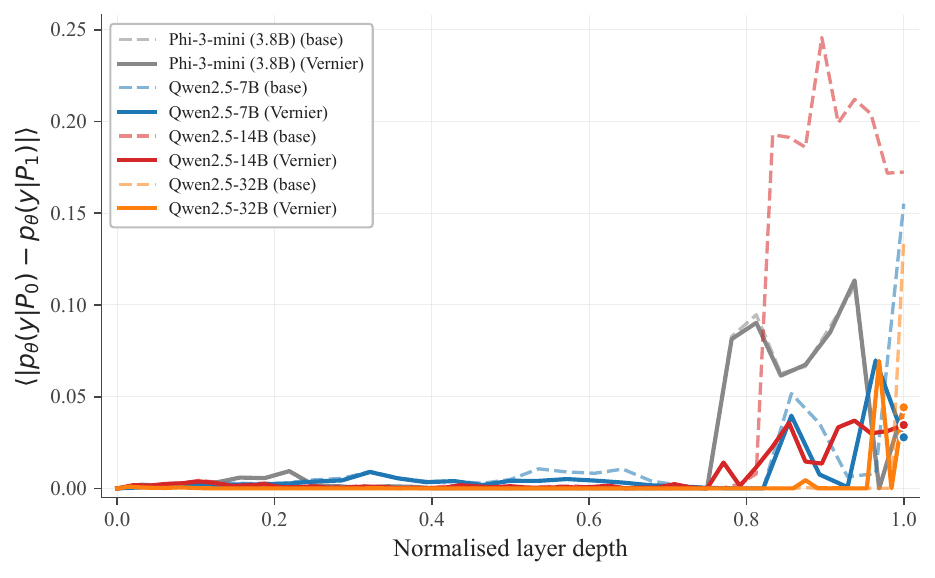}
  \end{minipage}\hfill
  \begin{minipage}[t]{0.48\linewidth}
    \centering
    \includegraphics[width=\linewidth]{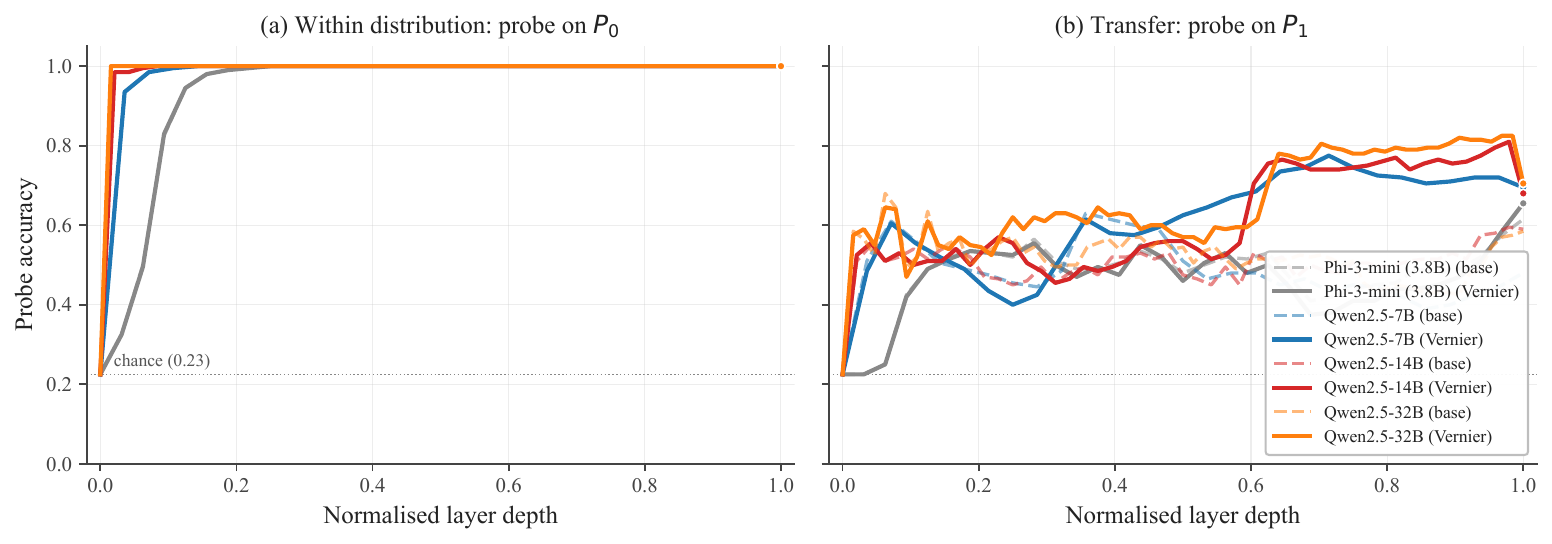}
  \end{minipage}
  \caption{Layer-wise mechanism diagnostics on held-out CLadder. (a) Last-token hidden-state cosine, (b) answer-token logit-lens disagreement, (c) and (d) variable-name probe accuracy. Dashed curves are base models, and solid curves are \method{} adapters.}
  \label{fig:mech}
\end{figure}

Two layer-wise measurements at the decision token are plotted for Phi-3 and the three Qwen scales (see \Cref{fig:mech}). The cosine similarity between the hidden states of \(\Pzero\) and \(\Pone\) measures whether the representation the model uses to predict depends on the surface form. The absolute logit-lens disagreement measures whether the model's intermediate belief about the answer depends on it. On all three Qwen scales \method{} raises the cosine and lowers the disagreement, with the effect concentrated in the upper third of the network. On Phi-3 neither measurement moves. The final-layer summaries across models are reported in \Cref{tab:mech}. The logit-lens reduction factor is 5.5 on Qwen-7B, 5.0 on Qwen-14B, and 3.1 on Qwen-32B, with the smaller 32B ratio explained by the lower base disagreement on Qwen-32B leaving less room to shrink. As a methodological control, we also computed the cosine pooled over all sequence positions rather than at the last non-pad token. The pooled signal moves by only 0.006 at the final layer on Qwen-7B against a 0.032 shift in the last-token cosine. Pooling washes out the effect because \(T\) edits only a small number of variable-name tokens.

The cosine result has a possible confound: \method{} might make the perturbed view content-free rather than aligned. We test that with a layer-wise linear probe (see \Cref{fig:mech,tab:mech}). At each layer, we train multinomial logistic regression to predict the variable-name family (12 classes: the 11 most frequent surface forms plus an ``other'' bucket) from the final prompt-token state of \(\Pzero\), then evaluate the same probe on \(\Pone\). If the update removed story content from \(\Pone\), transfer accuracy should fall toward chance. It rises instead. Final-layer transfer goes from 0.615 to 0.655 on Phi-3, 0.480 to 0.695 on Qwen-7B, 0.590 to 0.680 on Qwen-14B, and 0.585 to 0.705 on Qwen-32B. The adapted model recovers more of the same story-level information from the perturbed lexical form.

\paragraph{The alignment comes from paired views, while the KL sharpens beliefs.} To attribute the alignment correctly we run the same diagnostics on the augmentation baseline (\(\beta = 0\), no KL), reported in \Cref{tab:twinmech}. On Qwen-7B augmentation alone raises the probe transfer to 0.690, against 0.695 for the full \method{} loss, and the final-layer cosine is the same under augmentation and the full loss. On Qwen-14B augmentation gives a probe transfer of 0.730, above \method{}'s 0.680. The representational alignment is produced by training on both views, not by the consistency term alone. The one diagnostic on which the KL moves the needle further is the logit-lens disagreement, which augmentation brings from 0.155 to 0.074 and the KL halves again to 0.028 on Qwen-7B. The KL therefore refines the agreement of the intermediate answer beliefs without changing the recoverability of story content.

\begin{wrapfigure}{r}{0.6\textwidth}
  \centering
  \vspace{-15pt}
  \includegraphics[width=1.0\linewidth]{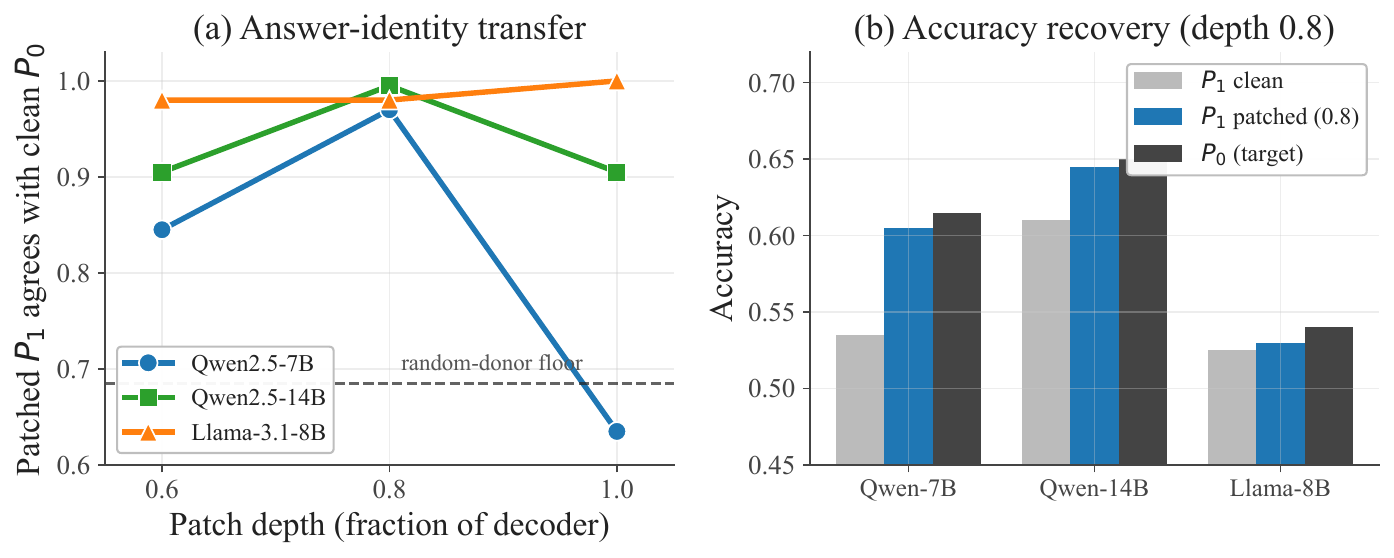}
  \caption{Causal activation patching on base Qwen2.5-7B, Qwen2.5-14B,
  and Llama-3.1-8B (held-out CLadder, $n = 200$). The \(\Pzero\)
  decision-token hidden state is copied into the corresponding \(\Pone\)
  state at the listed decoder depth. (a) Fraction of items on which the
  patched \(\Pone\) prediction agrees with the clean \(\Pzero\)
  prediction, peaking at depth 0.8; the dashed line is the random-donor
  floor (0.685) from \Cref{fig:patch-controls}. (b) Accuracy recovery at
  depth 0.8, where the patched \(\Pone\) moves from its clean value toward
  the \(\Pzero\) target. On the \method-adapted models the two views
  already agree, so the same patch is close to a no-op. The full per-depth
  and \method-adapted numbers are tabulated in \Cref{tab:patch}.}
  \label{fig:patch}
  \vspace{-15pt}
\end{wrapfigure}

\paragraph{A causal test by activation patching.}
The cosine, logit-lens, and probe diagnostics are correlational. To test whether the decision-token representation participates causally in the answer difference between the two views, we run an activation-patching experiment \citep{zhang2024patching} on Qwen-7B, Qwen-14B, and Llama-3.1-8B (see \Cref{fig:patch}). For each item we cache the \(\Pzero\) hidden state at the decision token at a chosen decoder depth, overwrite the corresponding \(\Pone\) hidden state with it, and let the patched activation propagate to the answer. On the base Qwen models the patch transfers \(\Pzero\)'s answer to \(\Pone\). Patching Qwen-7B at depth 0.8 raises \(\Pone\) accuracy from 0.535 to 0.605, close to \(\Pzero\)'s 0.615, and makes the patched \(\Pone\) prediction agree with \(\Pzero\) on 97 percent of items. The same depth-0.8 peak replicates on Qwen-14B, where the patch raises \(\Pone\) accuracy from 0.610 to 0.645 and the patched prediction agrees with \(\Pzero\) on 99.5 percent of items, against weaker effects at depth 0.6. Llama-3.1-8B has a smaller base gap on this CLadder slice, so the patch gives little accuracy lift, but it still makes patched \(\Pone\) agree with clean \(\Pzero\) on 98 to 100 percent of items. A larger 500-item depth-0.8 check gives the same agreement pattern, with bootstrap 95\% confidence intervals of 0.954--0.984 for Qwen-7B, 0.986--1.000 for Qwen-14B, and 0.962--0.988 for Llama-3.1-8B. Thus the patching result has two parts: accuracy recovery when the clean \(\Pzero\) answer is more accurate, and answer-identity transfer even when the clean gap is small. Donor controls on Qwen-7B show that this is not an arbitrary high-layer-vector effect: a different-prediction donor drives patched \(\Pone\) toward the donor answer and lowers accuracy (see \Cref{fig:patch-controls}). The controls also pin down the baseline against which the 97 percent agreement should be read. A random donor, which injects an unrelated item's \(\Pzero\) state, still leaves patched \(\Pone\) agreeing with the clean \(\Pzero\) prediction on only 0.685 of items, so the matched patch is a 0.285 lift over this uninformative-injection floor rather than a restatement of how often the two clean views already agree. The decision-token representation is therefore a causal control point for answer identity across two model families, with the accuracy benefit determined by whether \(\Pzero\) is the better view on that slice. On the \method-adapted models the two views already agree, so the same patch is close to a no-op, consistent with the alignment having already been learned.

\paragraph{The misalignment is item-conditional, not a global direction.} Patching uses the per-item \(\Pzero\) representation, which carries that item's content. This raises the question of whether the \(\Pzero - \Pone\) difference is instead a single direction, or a low-rank map, that could be estimated once and applied at inference with no fine-tuning. We test this directly. At the causal layer we fit an affine map \(T(h) = h + C_k h + b\) from \(\Pone\) to \(\Pzero\) decision-token states on 500 CLadder items, with \(b\) the mean difference and \(C_k\) a rank-\(k\) least-squares correction, and apply it to the \(\Pone\) decision token at inference over a rank sweep (see \Cref{tab:align}). The result is uniformly negative on Qwen-1.5B, 3B, 7B, 14B, and Phi-3. The mean-difference vector (\(k = 0\)) leaves the gap and \(\Pone\) accuracy unchanged, and any \(k \ge 4\) lowers \(\Pone\) accuracy rather than raising it. On Qwen-7B the held-out CLadder \(\Pone\) accuracy drops from 0.54 to 0.48 and the gap widens from 8.0 to 13.5 pp, and the CRASS and e-CARE gaps do not move. The failure is not specific to one layer or to a linear map. Sweeping the intervention over four decoder depths and replacing the affine map with a one-hidden-layer MLP fit on the same representations also fails to close the gap on Qwen-7B and Qwen-14B. The best case over this exhaustive sweep still leaves residual gaps of 5.0 pp on Qwen-7B CLadder, 31.0 pp on Qwen-7B CRASS, 4.5 pp on Qwen-14B CLadder, and 26.0 pp on Qwen-14B CRASS (see \Cref{tab:steer-exhaustive}). Several deeper configurations degrade \(\Pone\) further, for example a mid-depth MLP that drives the Qwen-14B CRASS gap from 27.5 to 59.0 pp. There is no consistent lexical-gap direction in the maps we tested. This sets the lexical gap apart from behavioural attributes such as sentiment, refusal, or truthfulness that prior work removes or controls with a single steering direction \citep{turner2023actadd, zou2023repe}. The decision-token difference appears dominated by item-specific content, so although the per-item representation controls answer identity, the tested global maps do not realign the views. This suggests that closure needs a weight update that relearns the read-out from each lexical form rather than an inference-time edit. The two views are not offset by a fixed transformation in our tests. They are read out by a function that the fine-tune repairs. This resistance is specific to the causal benchmarks. On the non-causal tasks a nonlinear map does partially realign the views.

\subsection{Sensitivity, robustness, and capability}
\label{sec:robust}

The auxiliary checks support the same scoped conclusion. The \(\beta\) sweep is stable around \(\beta\in[0.5,1]\), with over-regularisation visible at \(\beta=5\), and other hyperparameter sweeps leave the qualitative result unchanged (see \Cref{fig:appendix-diagnostics,tab:hyperparams}). Placeholder-scheme stress tests show that the learned invariance is not fully scheme-independent. Qwen-7B remains closed across held-out schemes, but Qwen-14B is brittle on \texttt{VAR\_A, VAR\_B} (see \Cref{tab:adversarial}). Larger-\(n\), paired, and stratified CLadder checks show that the in-distribution base gap is modest but the paired reduction remains positive on average (see \Cref{tab:stats,tab:large-cladder,tab:base-gap-significance,tab:stratified-splits}). Larger-OOD checks show a strong CRASS reduction on Qwen-7B and Qwen-14B but only a small e-CARE reduction (see \Cref{tab:ood-bign}). The sampled capability checks do not detect broad damage on the Qwen and Phi rows, but they are too small to certify capability preservation. Mistral is the clear failure case (see \Cref{fig:capability,fig:mistral-sweep}). Joint training on CLadder/CRASS/e-CARE trades a little CLadder closure for a much smaller CRASS gap (see \Cref{tab:settingB}). A lexical-overlap audit finds no 8-gram or 13-gram overlap between the CLadder training prompts and the sampled MMLU, HellaSwag, or GSM8K capability items (see \Cref{tab:contamination}), so the sampled capability changes are unlikely to be explained by verbatim train-test contamination. These checks narrow the claim rather than broaden it. The recipe is useful in a working regime, not a universally safe fine-tune.

\subsection{When the method works: capacity and base-model identity}
\label{sec:scale}

To turn the question of when \method{} works into a measured one, we add three sub-4B models and run the full diagnostic suite on each (see \Cref{tab:capacity,fig:appendix-diagnostics}). Two are from the Qwen family (Qwen2.5-1.5B and Qwen2.5-3B), so comparing them against the larger Qwen models isolates capacity with architecture and pretraining held fixed. The third is Gemma-2-2B, a second small family.

\textbf{A within-family capacity threshold.} Inside the Qwen family the success boundary lies between 1.5B and 3B by our joint criterion: the gap should close without \(\Pzero\) regression, probe transfer should rise, and logit-lens disagreement should fall. Both endpoints are measured over three seeds, so the boundary is not an artefact of a single run. Qwen-1.5B fails this joint criterion on all three seeds: after training the gap is \(+3.5 \pm 0.8\) pp, probe transfer drops by 0.09, and logit-lens disagreement barely moves. Qwen-3B succeeds on all three seeds by the same diagnostics: \(\Pzero\) and \(\Pone\) both rise to 0.88 with the gap at \(+0.5 \pm 0.7\) pp, the probe transfer rises by 0.29, the largest gain of any model, and the logit-lens disagreement halves. The two endpoints separate cleanly on the mechanism diagnostics and the per-seed spread (std under 1 pp on the gap) is far smaller than the gap between them. Since Qwen-1.5B and Qwen-3B share architecture and training family, this isolates capacity from base-model identity more cleanly than a cross-family comparison would.

\textbf{But scale is necessary only within a family, not sufficient across families.} The threshold is not a parameter-count rule. Phi-3-mini has 3.8B parameters and fails, while Qwen-3B is smaller and succeeds. Model family, pretraining, and instruction tuning also matter. Mistral-7B makes the same point at a larger scale because it needs a lower learning rate than the other 7B-class models. Gemma-2-2B is useful for a different reason. Its in-distribution gap closes with both views rising, but its logit-lens disagreement widens and probe transfer moves only 0.07. Gemma appears to fit the shared answer key without producing the representational alignment seen in the working models. Across the nine models, \method{} needs both enough representational headroom and a base model whose post-training leaves that headroom usable.

\textbf{The working regime (7B to 32B) is not Qwen-specific.} Qwen-7B, Qwen-14B, and Llama-3.1-8B agree across the behavioural diagnostics, and the closure reproduces across three seeds. Qwen-32B is harder to compare because its base in-distribution gap is already near zero, but the mechanism still moves: last-token cosine rises by 0.032, logit-lens disagreement drops 3.1-fold, and variable-name probe transfer rises by 0.12. e-CARE is the main exception. Its gap is within noise on Qwen-14B and reverses by 2.5 pp on Qwen-32B. We ran the same cosine diagnostic on e-CARE and found that \method{} still moves the two final-layer decision-token states closer together on all three Qwen scales (see \Cref{tab:ecare-mech}). The remaining e-CARE gap is plausibly downstream of this structural alignment, rather than simply a failure to align the final-layer representations.

\subsection{Preliminary non-causal rename checks}
\label{sec:generalise}

The account so far comes from causal question answering. We also run a preliminary check on two non-causal BBH tasks with object names replaced by typed placeholders: logical\_deduction and reasoning\_about\_colored\_objects (see \Cref{tab:logdeduct}). The logical\_deduction rows use 100 held-out items, while the colored-objects rows use a smaller 20-item diagnostic slice. Both have a base rename gap, 9.0 pp on Qwen-7B logical\_deduction and 20.0 to 45.0 pp on colored objects. Counterfactual augmentation closes or reduces the gap with both views rising rather than \(\Pzero\) collapsing. On colored objects the Qwen-7B gap goes from 20.0 to \(-5.0\) pp as \(\Pzero\) and \(\Pone\) rise from 0.500 and 0.300 to 0.800 and 0.850, and the Qwen-14B gap halves from 45.0 to 20.0 pp with both views rising. Activation patching reproduces the causal-sufficiency pattern on the tested non-causal cells. Patching the \(\Pzero\) decision-token state into \(\Pone\) raises logical\_deduction Qwen-7B \(\Pone\) accuracy from 0.63 to 0.77 with 89 percent agreement to \(\Pzero\), and raises colored-objects Qwen-14B \(\Pone\) accuracy from 0.25 to 0.75 with full agreement. On colored objects the Qwen-7B patch is present but peaks at a deeper layer rather than 0.8. These results suggest that the rename gap, closure by augmentation, and decision-token control-point pattern can extend beyond causal QA, but they should not be read as a full non-causal benchmark evaluation.

Steerability does differ by task. No inference-time map closes the gap on the causal benchmarks (\Cref{sec:mech}), but on both non-causal tasks a nonlinear map at the causal layer partially realigns the views. On Qwen-7B logical\_deduction, it reduces the gap from 9.0 to 1.0 pp and raises \(\Pone\) from 0.64 to 0.73. We tested whether the effective rank of the misalignment predicts this across five tasks and two model scales (see \Cref{fig:effrank}). It does not: CLadder has a lower effective rank than the steerable non-causal tasks but remains unsteerable, while colored objects has the highest effective ranks and is steerable on both Qwen scales. The source of this task dependence remains open.

\section{Discussion}
\label{sec:disc}

\paragraph{Alignment, not removal.}
\label{sec:disc-alignment}

The mechanism result is easy to misread. Higher cosine and lower logit-lens disagreement might look like removal of lexical-anchor information from the perturbed view. The probe says otherwise. After \method, the variable-name family is more recoverable from \(\Pone\) than before training, the opposite of what a simple information-removal account predicts. Activation patching turns this from a correlation into an intervention: overwriting the \(\Pone\) decision-token representation with the \(\Pzero\) representation transfers \(\Pzero\)'s answer identity to \(\Pone\) on 97 percent of Qwen-7B base-model items, 99.5 percent on Qwen-14B, and at least 98 percent on Llama-3.1-8B. The representation \method{} aligns is not a decorative diagnostic. It is a causal control point for the answer.

\paragraph{What the consistency term buys.}
\label{sec:disc-consistency}

\Cref{sec:ablation} and \Cref{sec:mech} leave a fairly simple decomposition. The paired lexical views are the behavioural engine. They close 87.5 percent of the in-distribution gap, give the CRASS transfer that survives at larger \(n\), and produce the same probe-transfer gain as the full loss (see \Cref{tab:twinmech}). The answer-subspace KL mostly sharpens intermediate answer beliefs. It halves the logit-lens disagreement beyond augmentation, but it does not improve the larger-\(n\) CRASS number and does not change the probe or cosine results. We keep the KL for its identifiability interpretation and belief-level effect, not because it drives the closure. R-Drop provides the negative control: consistency without the structure-preserving perturbation widens the gap.

\paragraph{Failure modes and limitations.}
\label{sec:limits}

The failures are informative. Phi-3 closes the in-distribution gap by \(\Pzero\) regression rather than \(\Pone\) improvement, and an MLP-targeted LoRA ablation reproduces the same shortcut (see \Cref{sec:phi3-mlp}). Mistral-7B needs a lower learning rate and still trades OOD transfer against capability (see \Cref{fig:capability,fig:mistral-sweep}). The largest model here is 32B under QLoRA, most held-out and capability cells use \(n=200\), the colored-objects non-causal check uses only \(n=20\), and the capacity threshold is clean only within Qwen. The capability table should be read as a screen for large degradation rather than as a standard full-benchmark evaluation. The mechanism claim is also bounded: the evidence argues against simple information erasure and against a fixed inference-time steering direction, but it does not prove that downstream computation is unchanged after \method{}. A learned read-out or decision-boundary recalibration conditioned on the item representation remains compatible with our misalignment account.

\section{Conclusion}
\label{sec:conclusion}

In the regimes tested here, the lexical gap is better explained by representational misalignment than by simple information loss. \method{} uses paired-view counterfactual augmentation to realign original and placeholder views, while the answer-subspace KL mainly refines answer-belief agreement. Probe transfer and activation patching show that the perturbed view retains recoverable story-level information and that the decision-token representation can carry answer identity across views. Fixed inference-time maps do not close the causal benchmarks, which suggests that the read-out repair is item-conditional and weight-mediated. The strongest transfer is on CRASS. e-CARE remains weak and task-dependent. The capacity and family sweep further bounds the claim. Qwen succeeds at 3B but not 1.5B, Phi-3 and Gemma expose failure modes, and Mistral requires a lower-rate recipe. Preliminary non-causal checks suggest that similar rename gaps can occur beyond causal QA, but a full account of those settings remains open.

\bibliographystyle{plainnat}
\bibliography{main}

@misc{caliper,
  title         = {{C}aliper: Probing Lexical Anchors versus Causal Structure in {LLM}s},
  author        = {Yu, Zhenyu and Zhou, Shuigeng},
  year          = {2026},
  eprint        = {2606.04915},
  archivePrefix = {arXiv},
  primaryClass  = {cs.CL},
  url           = {https://arxiv.org/abs/2606.04915}
}

@inproceedings{chi2025causalmirage,
  title         = {Unveiling Causal Reasoning in Large Language Models: Reality or Mirage?},
  author        = {Chi, Haoang and Li, He and Yang, Wenjing and Liu, Feng and Lan, Long
                   and Ren, Xiaoguang and Liu, Tongliang and Han, Bo},
  booktitle     = {Advances in Neural Information Processing Systems},
  year          = {2024}
}

@misc{hao2025mathrobust,
  title         = {An Investigation of Robustness of {LLM}s in Mathematical Reasoning:
                   Benchmarking with Mathematically-Equivalent Transformation of
                   Advanced Mathematical Problems},
  author        = {Hao, Yuren and Wan, Xiang and Zhai, ChengXiang},
  year          = {2025},
  note          = {arXiv:2508.08833}
}

@misc{lee2025econcausal,
  title         = {{E}con{C}ausal: A Context-Aware Economic Reasoning Benchmark for
                   Large Language Models},
  author        = {Lee, Donggyu and Yun, Hyeok and Cha, Meeyoung and Park, Sungwon
                   and Park, Sangyoon and Kim, Jihee},
  year          = {2025},
  note          = {arXiv:2510.07231}
}

@inproceedings{zhang2024patching,
  title         = {Towards Best Practices of Activation Patching in Language Models:
                   Metrics and Methods},
  author        = {Zhang, Fred and Nanda, Neel},
  booktitle     = {International Conference on Learning Representations (ICLR)},
  year          = {2024}
}

@inproceedings{cladder,
  title     = {{CL}adder: Assessing Causal Reasoning in Language Models},
  author    = {Jin, Zhijing and Chen, Yuen and Leeb, Felix and Gresele, Luigi
               and Kamal, Ojasv and Lyu, Zhiheng and Blin, Kevin and
               Adauto, Fernando Gonzalez and Kleiman-Weiner, Max and
               Sachan, Mrinmaya and Sch{\"o}lkopf, Bernhard},
  booktitle = {Advances in Neural Information Processing Systems},
  year      = {2023}
}

@inproceedings{crass,
  title     = {{CRASS}: A Novel Data Set and Benchmark to Test Counterfactual
               Reasoning of Large Language Models},
  author    = {Frohberg, J{\"o}rn and Binder, Frank},
  booktitle = {Proceedings of the 13th Language Resources and Evaluation
               Conference (LREC)},
  year      = {2022}
}

@inproceedings{bbh,
  title     = {Challenging {BIG}-Bench Tasks and Whether Chain-of-Thought
               Can Solve Them},
  author    = {Suzgun, Mirac and Scales, Nathan and Sch{\"a}rli, Nathanael
               and Gehrmann, Sebastian and Tay, Yi and Chung, Hyung Won
               and Chowdhery, Aakanksha and Le, Quoc V. and Chi, Ed H.
               and Zhou, Denny and Wei, Jason},
  booktitle = {Findings of the Association for Computational Linguistics: ACL 2023},
  year      = {2023}
}

@inproceedings{ecare,
  title     = {e-{CARE}: a New Dataset for Exploring Explainable Causal Reasoning},
  author    = {Du, Li and Ding, Xiao and Xiong, Kai and Liu, Ting and Qin, Bing},
  booktitle = {Proceedings of the 60th Annual Meeting of the Association
               for Computational Linguistics (ACL)},
  year      = {2022}
}

@book{pearl2009,
  author    = {Pearl, Judea},
  title     = {Causality: Models, Reasoning, and Inference},
  publisher = {Cambridge University Press},
  edition   = {2nd},
  year      = {2009}
}

@book{pearl2018,
  author    = {Pearl, Judea and Mackenzie, Dana},
  title     = {The Book of Why: The New Science of Cause and Effect},
  publisher = {Basic Books},
  year      = {2018}
}

@inproceedings{wei2022cot,
  title     = {Chain-of-Thought Prompting Elicits Reasoning in Large Language Models},
  author    = {Wei, Jason and Wang, Xuezhi and Schuurmans, Dale and Bosma, Maarten
               and Ichter, Brian and Xia, Fei and Chi, Ed and Le, Quoc V and Zhou, Denny},
  booktitle = {Advances in Neural Information Processing Systems},
  year      = {2022}
}

@inproceedings{kojima2022zeroshot,
  title     = {Large Language Models are Zero-Shot Reasoners},
  author    = {Kojima, Takeshi and Gu, Shixiang Shane and Reid, Machel and
               Matsuo, Yutaka and Iwasawa, Yusuke},
  booktitle = {Advances in Neural Information Processing Systems},
  year      = {2022}
}

@inproceedings{wang2023selfconsistency,
  title     = {Self-Consistency Improves Chain of Thought Reasoning in Language Models},
  author    = {Wang, Xuezhi and Wei, Jason and Schuurmans, Dale and Le, Quoc and
               Chi, Ed and Narang, Sharan and Chowdhery, Aakanksha and Zhou, Denny},
  booktitle = {International Conference on Learning Representations},
  year      = {2023}
}

@inproceedings{ouyang2022instructgpt,
  title     = {Training language models to follow instructions with human feedback},
  author    = {Ouyang, Long and Wu, Jeffrey and Jiang, Xu and Almeida, Diogo and
               Wainwright, Carroll L and Mishkin, Pamela and Zhang, Chong and
               Agarwal, Sandhini and Slama, Katarina and Ray, Alex and others},
  booktitle = {Advances in Neural Information Processing Systems},
  year      = {2022}
}

@inproceedings{sanh2022t0,
  title     = {Multitask Prompted Training Enables Zero-Shot Task Generalization},
  author    = {Sanh, Victor and Webson, Albert and Raffel, Colin and Bach, Stephen H
               and Sutawika, Lintang and Alyafeai, Zaid and Chaffin, Antoine and
               Stiegler, Arnaud and Le Scao, Teven and Raja, Arun and others},
  booktitle = {International Conference on Learning Representations},
  year      = {2022}
}

@inproceedings{wei2022flan,
  title     = {Finetuned language models are zero-shot learners},
  author    = {Wei, Jason and Bosma, Maarten and Zhao, Vincent Y and Guu, Kelvin
               and Yu, Adams Wei and Lester, Brian and Du, Nan and Dai, Andrew M
               and Le, Quoc V},
  booktitle = {International Conference on Learning Representations},
  year      = {2022}
}

@inproceedings{mccoy2019hans,
  title     = {Right for the Wrong Reasons: Diagnosing Syntactic Heuristics in
               Natural Language Inference},
  author    = {McCoy, Tom and Pavlick, Ellie and Linzen, Tal},
  booktitle = {Proceedings of the 57th Annual Meeting of the Association
               for Computational Linguistics (ACL)},
  year      = {2019}
}

@inproceedings{webson2022prompts,
  title     = {Do Prompt-Based Models Really Understand the Meaning of Their Prompts?},
  author    = {Webson, Albert and Pavlick, Ellie},
  booktitle = {Proceedings of the 2022 Conference of the North American Chapter
               of the Association for Computational Linguistics (NAACL)},
  year      = {2022}
}

@inproceedings{sclar2024prompt,
  title     = {Quantifying Language Models' Sensitivity to Spurious Features
               in Prompt Design or: How I learned to start worrying about
               prompt formatting},
  author    = {Sclar, Melanie and Choi, Yejin and Tsvetkov, Yulia and Suhr, Alane},
  booktitle = {International Conference on Learning Representations},
  year      = {2024}
}

@inproceedings{kaushik2020learning,
  title     = {Learning the Difference that Makes a Difference with
               Counterfactually-Augmented Data},
  author    = {Kaushik, Divyansh and Hovy, Eduard and Lipton, Zachary C.},
  booktitle = {International Conference on Learning Representations},
  year      = {2020}
}

@inproceedings{min2022rethinking,
  title     = {Rethinking the Role of Demonstrations: What Makes In-Context Learning Work?},
  author    = {Min, Sewon and Lyu, Xinxi and Holtzman, Ari and Artetxe, Mikel
               and Lewis, Mike and Hajishirzi, Hannaneh and Zettlemoyer, Luke},
  booktitle = {Proceedings of the 2022 Conference on Empirical Methods
               in Natural Language Processing (EMNLP)},
  year      = {2022}
}

@inproceedings{rdrop,
  title     = {{R}-Drop: Regularized Dropout for Neural Networks},
  author    = {Liang, Xiaobo and Wu, Lijun and Li, Juntao and Wang, Yue
               and Meng, Qi and Qin, Tao and Chen, Wei and Zhang, Min
               and Liu, Tie-Yan},
  booktitle = {Advances in Neural Information Processing Systems},
  year      = {2021}
}

@inproceedings{simcse,
  title     = {{S}im{CSE}: Simple Contrastive Learning of Sentence Embeddings},
  author    = {Gao, Tianyu and Yao, Xingcheng and Chen, Danqi},
  booktitle = {Proceedings of the 2021 Conference on Empirical Methods
               in Natural Language Processing (EMNLP)},
  year      = {2021}
}

@misc{elhage2021circuits,
  title  = {A Mathematical Framework for Transformer Circuits},
  author = {Elhage, Nelson and Nanda, Neel and Olsson, Catherine and Henighan, Tom
            and Joseph, Nicholas and Mann, Ben and Askell, Amanda and Bai, Yuntao
            and Chen, Anna and Conerly, Tom and others},
  year   = {2021},
  note   = {Transformer Circuits Thread.}
}

@inproceedings{meng2022rome,
  title     = {Locating and Editing Factual Associations in {GPT}},
  author    = {Meng, Kevin and Bau, David and Andonian, Alex and Belinkov, Yonatan},
  booktitle = {Advances in Neural Information Processing Systems},
  year      = {2022}
}

@inproceedings{geva2023dissecting,
  title     = {Dissecting Recall of Factual Associations in Auto-Regressive
               Language Models},
  author    = {Geva, Mor and Bastings, Jasmijn and Filippova, Katja and Globerson, Amir},
  booktitle = {Proceedings of the 2023 Conference on Empirical Methods
               in Natural Language Processing (EMNLP)},
  year      = {2023}
}

@misc{belrose2023tunedlens,
  title  = {Eliciting Latent Predictions from Transformers with the Tuned Lens},
  author = {Belrose, Nora and Ostrovsky, Igor and McKinney, Lev and Furman, Zach
            and Smith, Logan and Halawi, Danny and Biderman, Stella and
            Steinhardt, Jacob},
  year   = {2023},
  note   = {arXiv:2303.08112}
}

@misc{turner2023actadd,
  title  = {Activation Addition: Steering Language Models Without Optimization},
  author = {Turner, Alexander Matt and Thiergart, Lisa and Leech, Gavin and
            Udell, David and Mini, Ulisse and MacDiarmid, Monte},
  year   = {2023},
  note   = {arXiv:2308.10248}
}

@misc{zou2023repe,
  title  = {Representation Engineering: A Top-Down Approach to {AI} Transparency},
  author = {Zou, Andy and Phan, Long and Chen, Sarah and Campbell, James
            and Guo, Phillip and Ren, Richard and others},
  year   = {2023},
  note   = {arXiv:2310.01405}
}

@inproceedings{lora,
  title     = {{L}o{RA}: Low-Rank Adaptation of Large Language Models},
  author    = {Hu, Edward J. and Shen, Yelong and Wallis, Phillip and
               Allen-Zhu, Zeyuan and Li, Yuanzhi and Wang, Shean and
               Wang, Lu and Chen, Weizhu},
  booktitle = {International Conference on Learning Representations (ICLR)},
  year      = {2022}
}

@inproceedings{qlora,
  title     = {{QL}o{RA}: Efficient Finetuning of Quantized {LLM}s},
  author    = {Dettmers, Tim and Pagnoni, Artidoro and Holtzman, Ari
               and Zettlemoyer, Luke},
  booktitle = {Advances in Neural Information Processing Systems},
  year      = {2023}
}

@inproceedings{adamw,
  title     = {Decoupled Weight Decay Regularization},
  author    = {Loshchilov, Ilya and Hutter, Frank},
  booktitle = {International Conference on Learning Representations (ICLR)},
  year      = {2019}
}

@inproceedings{mmlu,
  title     = {Measuring Massive Multitask Language Understanding},
  author    = {Hendrycks, Dan and Burns, Collin and Basart, Steven and
               Zou, Andy and Mazeika, Mantas and Song, Dawn and Steinhardt, Jacob},
  booktitle = {International Conference on Learning Representations (ICLR)},
  year      = {2021}
}

@misc{phi3,
  title  = {Phi-3 Technical Report: A Highly Capable Language Model
            Locally on Your Phone},
  author = {Abdin, Marah and Aneja, Jyoti and Awadalla, Hany and others},
  year   = {2024},
  note   = {arXiv:2404.14219}
}

@misc{qwen25,
  title  = {{Q}wen2.5 Technical Report},
  author = {{Qwen Team, Alibaba Group}},
  year   = {2024},
  note   = {arXiv:2412.15115}
}

@misc{llama3,
  title  = {The Llama 3 Herd of Models},
  author = {{Llama Team, AI @ Meta}},
  year   = {2024},
  note   = {arXiv:2407.21783}
}

@misc{jiang2023mistral,
  title  = {Mistral 7{B}},
  author = {Jiang, Albert Q. and Sablayrolles, Alexandre and Mensch, Arthur
            and others},
  year   = {2023},
  note   = {arXiv:2310.06825}
}

@misc{gemma2,
  title  = {Gemma 2: Improving Open Language Models at a Practical Size},
  author = {{Gemma Team}},
  year   = {2024},
  note   = {arXiv:2408.00118}
}

@misc{wang2026causalflip,
  title         = {CausalFlip: A Benchmark for LLM Causal Judgment Beyond Semantic Matching},
  author        = {Wang, Yuzhe and Zhu, Yaochen and Li, Jundong},
  year          = {2026},
  eprint        = {2602.20094},
  archivePrefix = {arXiv},
  primaryClass  = {cs.CL},
  url           = {https://arxiv.org/abs/2602.20094}
}

@misc{li2026meter,
  title         = {METER: Evaluating Multi-Level Contextual Causal Reasoning in Large Language Models},
  author        = {Li, Pengfeng and Huang, Chen and Hao, Chaoqun and Chen, Hongyao and Wei, Xiao-Yong and Lei, Wenqiang and Ng, See-Kiong},
  year          = {2026},
  eprint        = {2604.11502},
  archivePrefix = {arXiv},
  primaryClass  = {cs.CL},
  url           = {https://arxiv.org/abs/2604.11502}
}

@inproceedings{cohen2016group,
  title     = {Group Equivariant Convolutional Networks},
  author    = {Cohen, Taco and Welling, Max},
  booktitle = {International Conference on Machine Learning (ICML)},
  year      = {2016}
}

@inproceedings{xie2020uda,
  title     = {Unsupervised Data Augmentation for Consistency Training},
  author    = {Xie, Qizhe and Dai, Zihang and Hovy, Eduard and Luong, Thang
               and Le, Quoc},
  booktitle = {Advances in Neural Information Processing Systems (NeurIPS)},
  year      = {2020}
}

\appendix
\setcounter{figure}{0}
\setcounter{table}{0}
\renewcommand{\thefigure}{A\arabic{figure}}
\renewcommand{\thetable}{A\arabic{table}}
\renewcommand{\theHfigure}{A\arabic{figure}}
\renewcommand{\theHtable}{A\arabic{table}}

\section{Additional diagnostic figures}
\label{sec:additional-figures}

\begin{figure}[H]
  \centering
  \begin{minipage}[t]{0.58\linewidth}
    \centering
    \includegraphics[width=\linewidth]{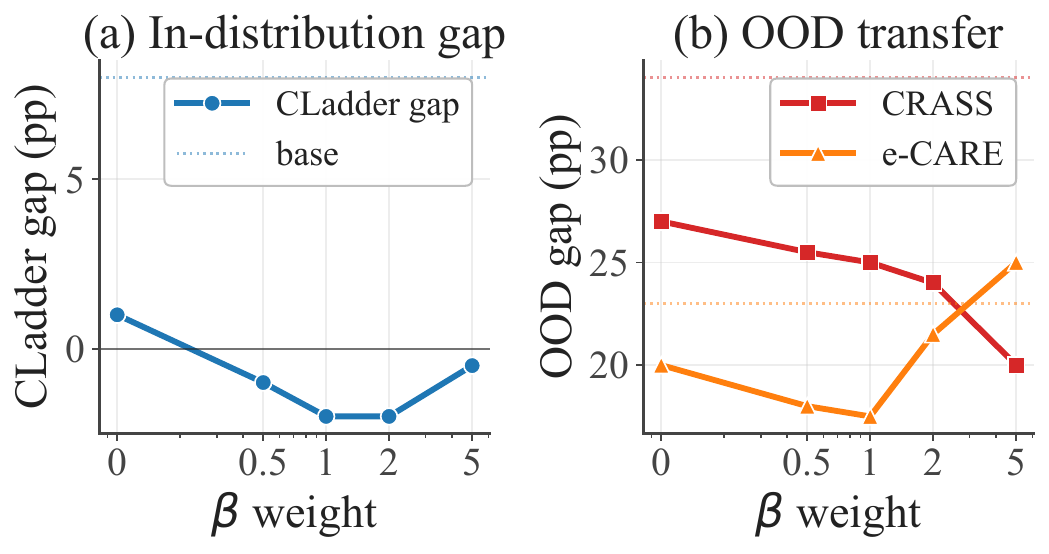}
  \end{minipage}\hfill
  \begin{minipage}[t]{0.38\linewidth}
    \centering
    \includegraphics[width=\linewidth]{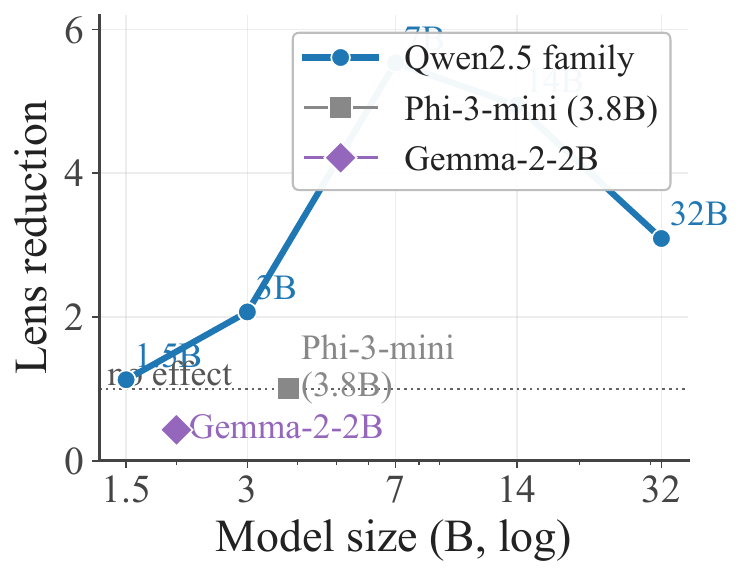}
  \end{minipage}
  \caption{Additional diagnostics. The left panel gives the Qwen2.5-7B sweep of the consistency-loss weight \(\beta\). The right panel gives logit-lens reduction by scale.}
  \label{fig:appendix-diagnostics}
\end{figure}

The consistency-weight sweep and the scale summary for the logit-lens reduction diagnostic are reported in \Cref{fig:appendix-diagnostics}. The stable region around \(\beta\in[0.5,1]\) supports the default choice in the main text, while the degradation at \(\beta=5\) marks the point where the consistency penalty starts to dominate task supervision. The scale panel is kept in the appendix because the model-by-model interpretation is already reported in the main text.

\section{Additional results tables and figures}
\label{sec:additional-tables}

\begin{table}[H]
\centering
\caption{Loss-component ablation on Qwen2.5-7B (held-out CLadder
$n=200$, CRASS $n=200$, e-CARE $n=200$), all at step 500 with
seed 42. Best results in each column are in bold.}
\label{tab:ablation}
\small
\setlength{\tabcolsep}{4pt}
\resizebox{\textwidth}{!}{%
\begin{tabular}{l ccc cc c}
\toprule
& \multicolumn{3}{c}{\textbf{In-distribution (CLadder)}}
& \multicolumn{2}{c}{\textbf{Out-of-distribution}} & \\
\cmidrule(lr){2-4} \cmidrule(lr){5-6}
\textbf{Variant} & $P_0$ & $P_1$ & \textbf{Gap}
        & \textbf{CRASS Gap} & \textbf{e-CARE Gap}
        & \textbf{Mean $\Delta_{\text{OOD}}$} \\
\midrule
Base (no fine-tuning)
   & 0.625 & 0.545 & $+8.0$
   & $+34.0$ & $+23.0$ & --- \\
$P_0$-only SFT
   & 0.850 & 0.755 & $+9.5$
   & $+28.0\;(\text{-}18\%)$ & $+23.5\;(\text{-}0\%)$ & $-9\%$ \\
R-Drop \citep{rdrop} (no $T$)
   & 0.965 & 0.780 & $+18.5$
   & $+25.5\;(\text{-}25\%)$ & $+22.0\;(\text{-}4\%)$ & $-15\%$ \\
Twin task ($\beta=0$, no KL)
   & 0.885 & 0.875 & $+1.0\;(\text{-}87.5\%)$
   & $+27.0\;(\text{-}21\%)$ & $+20.0\;(\text{-}13\%)$ & $-17\%$ \\
Full-vocab KL ($\beta=1$)
   & 0.930 & 0.925 & $+0.5\;(\text{-}94\%)$
   & $\mathbf{+21.5\;(\text{-}37\%)}$ & $+21.5\;(\text{-}7\%)$ & $-22\%$ \\
\textbf{Vernier} ($\beta=1$, answer-subspace KL)
   & 0.870 & 0.890 & $\mathbf{-2.0\;(\text{-}125\%)}$
   & $+25.0\;(\text{-}26\%)$
   & $\mathbf{+17.5\;(\text{-}24\%)}$
   & $\mathbf{-25\%}$ \\
\bottomrule
\end{tabular}
}
\end{table}

\begin{table}[H]
\centering
\caption{Counterfactual augmentation versus the full answer-subspace
KL loss on CRASS. Both adapters are trained only on CLadder. The
274-item rows use all prepared CRASS items.}
\label{tab:crass-aug-kl}
\small
\setlength{\tabcolsep}{5pt}
\begin{tabular}{l r l rr r r}
\toprule
\textbf{Model} & \(n\) & \textbf{Setting} & \(P_0\) & \(P_1\) &
\textbf{Gap} & \textbf{\(\Delta\) gap} \\
\midrule
\multirow{3}{*}{Qwen2.5-7B} & \multirow{3}{*}{200}
  & Base      & 0.845 & 0.505 & \(+34.0\) & -- \\
  & & Twin task & 0.890 & 0.650 & \(+24.0\) & \(-29\%\) \\
  & & Vernier   & 0.880 & 0.635 & \(+24.5\) & \(-28\%\) \\
\midrule
\multirow{3}{*}{Qwen2.5-7B} & \multirow{3}{*}{274}
  & Base      & 0.839 & 0.496 & \(+34.3\) & -- \\
  & & Twin task & 0.891 & 0.708 & \(\mathbf{+18.2}\) & \(\mathbf{-47\%}\) \\
  & & Vernier   & 0.872 & 0.639 & \(+23.4\) & \(-32\%\) \\
\midrule
\multirow{3}{*}{Qwen2.5-14B} & \multirow{3}{*}{200}
  & Base      & 0.880 & 0.610 & \(+27.0\) & -- \\
  & & Twin task & 0.915 & 0.715 & \(+20.0\) & \(-26\%\) \\
  & & Vernier   & 0.920 & 0.720 & \(+20.0\) & \(-26\%\) \\
\midrule
\multirow{3}{*}{Qwen2.5-14B} & \multirow{3}{*}{274}
  & Base      & 0.894 & 0.617 & \(+27.7\) & -- \\
  & & Twin task & 0.920 & 0.745 & \(+17.5\) & \(-37\%\) \\
  & & Vernier   & 0.923 & 0.759 & \(\mathbf{+16.4}\) & \(\mathbf{-41\%}\) \\
\bottomrule
\end{tabular}
\end{table}

\begin{table}[H]
\centering
\caption{Final-layer mechanism diagnostics across models, base versus
\method. Cosine is last-token hidden-state cosine
$\cos(h_L(\Pzero), h_L(\Pone))$ at final layer $L$. Logit-lens
disagreement is
$\langle |p_\theta(y \mid \Pzero) - p_\theta(y \mid \Pone)|\rangle$.
Probe transfer is linear-probe accuracy on final-layer $P_1$ hidden
states.}
\label{tab:mech}
\small
\setlength{\tabcolsep}{6pt}
\resizebox{\textwidth}{!}{%
\begin{tabular}{l rrr rrr rrr}
\toprule
& \multicolumn{3}{c}{\textbf{Hidden cosine}}
& \multicolumn{3}{c}{\textbf{Logit-lens gap}}
& \multicolumn{3}{c}{\textbf{Probe transfer}} \\
\cmidrule(lr){2-4} \cmidrule(lr){5-7} \cmidrule(lr){8-10}
\textbf{Model} & Base & Vernier & $\Delta$ & Base & Vernier & Ratio & Base & Vernier & $\Delta$ \\
\midrule
Phi-3-mini (3.8B) & $0.990$ & $0.990$ & $0.000$
                  & $0.044$ & $0.044$ & $1.0\times$
                  & $0.615$ & $0.655$ & $+0.040$ \\
Qwen2.5-7B        & $0.963$ & $\mathbf{0.995}$ & $\mathbf{+0.032}$
                  & $0.155$ & $\mathbf{0.028}$ & $\mathbf{5.5\times}$
                  & $0.480$ & $\mathbf{0.695}$ & $\mathbf{+0.215}$ \\
Qwen2.5-14B       & $0.979$ & $0.995$ & $+0.016$
                  & $0.173$ & $0.035$ & $5.0\times$
                  & $0.590$ & $0.680$ & $+0.090$ \\
Qwen2.5-32B       & $0.960$ & $0.992$ & $+0.032$
                  & $0.136$ & $0.044$ & $3.1\times$
                  & $0.585$ & $0.705$ & $+0.120$ \\
\bottomrule
\end{tabular}
}
\end{table}

\begin{table}[H]
\centering
\caption{Mechanism diagnostics for augmentation without KL and full
\method{} (held-out CLadder, $n = 200$). Augment is the $\beta = 0$
twin-task adapter. Probe transfer is final-layer variable-name probe
accuracy on $P_1$ hidden states, and logit-lens gap is final-layer
answer-token disagreement.}
\label{tab:twinmech}
\small
\setlength{\tabcolsep}{8pt}
\begin{tabular}{l l rrr}
\toprule
\textbf{Model} & \textbf{Diagnostic} & \textbf{Base} & \textbf{Augment ($\beta{=}0$)} & \textbf{Vernier ($\beta{=}1$)} \\
\midrule
\multirow{2}{*}{Qwen2.5-7B}
  & Probe transfer ($\uparrow$)   & 0.480 & $\mathbf{0.690}$ & 0.695 \\
  & Logit-lens gap ($\downarrow$) & 0.155 & 0.074 & $\mathbf{0.028}$ \\
\midrule
Qwen2.5-14B
  & Probe transfer ($\uparrow$)   & 0.590 & $\mathbf{0.730}$ & 0.680 \\
\bottomrule
\end{tabular}
\end{table}

\begin{table}[H]
\centering
\caption{Training-free linear realignment at inference. An affine map
$T(h) = h + C_k h + b$ is fit from $P_1$ to $P_0$ decision-token states
at depth 0.8 on 500 CLadder items and applied to the $P_1$ decision
token on the base model. Rank 8 is shown as representative.}
\label{tab:align}
\small
\setlength{\tabcolsep}{6pt}
\begin{tabular}{l rr rr rr}
\toprule
& \multicolumn{2}{c}{\textbf{CLadder gap}}
& \multicolumn{2}{c}{\textbf{CLadder $P_1$}}
& \multicolumn{2}{c}{\textbf{CRASS gap}} \\
\cmidrule(lr){2-3} \cmidrule(lr){4-5} \cmidrule(lr){6-7}
\textbf{Model} & clean & aligned & clean & aligned & clean & aligned \\
\midrule
Qwen2.5-1.5B & $-10.5$ & $-9.0$  & 0.53 & 0.52 & $+21.5$ & $+22.5$ \\
Qwen2.5-3B   & $-1.5$  & $-0.5$  & 0.53 & 0.52 & $+32.0$ & $+32.0$ \\
Qwen2.5-7B   & $+8.0$  & $+13.5$ & 0.54 & 0.48 & $+33.5$ & $+34.0$ \\
Qwen2.5-14B  & $+4.5$  & $+16.5$ & 0.60 & 0.48 & $+27.5$ & $+26.5$ \\
Phi-3-mini   & $+3.0$  & $+9.5$  & 0.56 & 0.48 & $+28.5$ & $+23.5$ \\
\bottomrule
\end{tabular}
\end{table}

\begin{table}[H]
\centering
\caption{Best result from exhaustive inference-time steering on causal
QA. Each row selects the smallest residual gap over four decoder depths
and both linear and MLP maps fitted from \(P_1\) to \(P_0\)
decision-token states.}
\label{tab:steer-exhaustive}
\small
\setlength{\tabcolsep}{6pt}
\begin{tabular}{l l l rr rr}
\toprule
\textbf{Model} & \textbf{Benchmark} & \textbf{Best map}
& \textbf{clean gap} & \textbf{aligned gap}
& \textbf{\(P_1\)} & \textbf{\(P_1\) aligned} \\
\midrule
Qwen2.5-7B  & CLadder & MLP@0.8    & \(+8.0\)  & \(+5.0\)  & 0.535 & 0.565 \\
Qwen2.5-7B  & CRASS   & linear@0.4 & \(+33.5\) & \(+31.0\) & 0.510 & 0.535 \\
Qwen2.5-14B & CLadder & MLP@0.6    & \(+4.5\)  & \(+4.5\)  & 0.605 & 0.605 \\
Qwen2.5-14B & CRASS   & linear@0.6 & \(+27.5\) & \(+26.0\) & 0.605 & 0.620 \\
\bottomrule
\end{tabular}
\end{table}

\begin{table}[t]
\centering
\caption{Causal activation patching on Qwen2.5-7B, Qwen2.5-14B, and
Llama-3.1-8B
(held-out CLadder, $n = 200$). The $P_0$ decision-token hidden state at
the listed decoder depth is copied into the corresponding $P_1$ state.
``$P_1$ patched'' is post-patch $P_1$ accuracy, and ``flip-to-$P_0$'' is
agreement with the clean $P_0$ prediction.}
\label{tab:patch}
\small
\setlength{\tabcolsep}{6pt}
\begin{tabular}{l l r rrr r}
\toprule
\textbf{Model} & \textbf{Setting} & \textbf{Depth} & \textbf{$P_0$} & \textbf{$P_1$ clean} & \textbf{$P_1$ patched} & \textbf{flip-to-$P_0$} \\
\midrule
\multirow{6}{*}{Qwen2.5-7B}
  & \multirow{3}{*}{Base}
    & 0.6 & 0.615 & 0.535 & 0.540 & 0.845 \\
  & & 0.8 & 0.615 & 0.535 & $\mathbf{0.605}$ & $\mathbf{0.970}$ \\
  & & 1.0 & 0.615 & 0.535 & 0.590 & 0.635 \\
\cmidrule(lr){2-7}
  & \multirow{3}{*}{Vernier}
    & 0.6 & 0.870 & 0.890 & 0.890 & 0.920 \\
  & & 0.8 & 0.870 & 0.890 & 0.870 & 1.000 \\
  & & 1.0 & 0.870 & 0.890 & 0.865 & 0.985 \\
\midrule
\multirow{6}{*}{Qwen2.5-14B}
  & \multirow{3}{*}{Base}
    & 0.6 & 0.650 & 0.610 & 0.625 & 0.905 \\
  & & 0.8 & 0.650 & 0.610 & $\mathbf{0.645}$ & $\mathbf{0.995}$ \\
  & & 1.0 & 0.650 & 0.610 & 0.645 & 0.905 \\
\cmidrule(lr){2-7}
  & \multirow{3}{*}{Vernier}
    & 0.6 & 0.910 & 0.935 & 0.935 & 0.955 \\
  & & 0.8 & 0.910 & 0.935 & 0.915 & 0.995 \\
  & & 1.0 & 0.910 & 0.935 & 0.885 & 0.925 \\
\midrule
\multirow{6}{*}{Llama-3.1-8B}
  & \multirow{3}{*}{Base}
    & 0.6 & 0.540 & 0.525 & 0.530 & 0.980 \\
  & & 0.8 & 0.540 & 0.525 & 0.530 & 0.980 \\
  & & 1.0 & 0.540 & 0.525 & 0.540 & 1.000 \\
\cmidrule(lr){2-7}
  & \multirow{3}{*}{Vernier}
    & 0.6 & 0.945 & 0.960 & 0.945 & 1.000 \\
  & & 0.8 & 0.945 & 0.960 & 0.945 & 1.000 \\
  & & 1.0 & 0.945 & 0.960 & 0.945 & 0.980 \\
\bottomrule
\end{tabular}
\end{table}

\begin{figure}[H]
  \centering
  \includegraphics[width=0.96\linewidth]{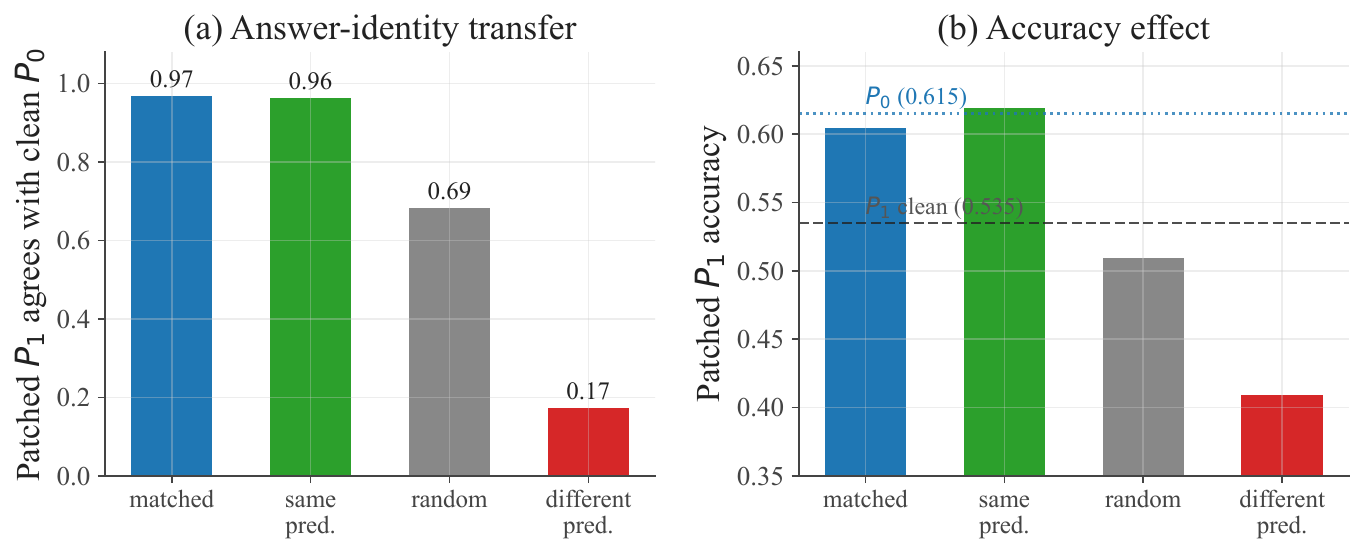}
  \caption{Activation-patching donor controls on base Qwen2.5-7B
  (held-out CLadder, \(n = 200\), depth 0.8). ``Matched'' copies the same
  item's \(\Pzero\) hidden state, and the controls copy another item's
  \(\Pzero\) state. (a) Agreement of the patched \(\Pone\) prediction with
  the clean \(\Pzero\) prediction. (b) Patched \(\Pone\) accuracy, with the
  clean-\(\Pone\) (0.535) and \(\Pzero\) (0.615) references. The matched
  and same-prediction donors transfer answer identity, the random donor
  sits at an uninformative 0.685 floor, and a different-prediction donor
  drives \(\Pone\) toward the wrong answer. The exact values are in
  \Cref{tab:patch-controls}.}
  \label{fig:patch-controls}
\end{figure}
\begin{table}[H]
\centering
\caption{Activation-patching donor controls on Qwen2.5-7B
(held-out CLadder, \(n=200\), depth 0.8). Matched uses the same item's
\(P_0\) hidden state, and the controls replace it with another item's
\(P_0\) hidden state.}
\label{tab:patch-controls}
\small
\setlength{\tabcolsep}{6pt}
\begin{tabular}{l l rrr rr}
\toprule
\textbf{Setting} & \textbf{Donor} & \textbf{\(P_0\)}
& \textbf{\(P_1\) clean} & \textbf{\(P_1\) patched}
& \textbf{flip target} & \textbf{flip donor} \\
\midrule
Base & matched & 0.615 & 0.535 & 0.605 & 0.970 & 0.970 \\
Base & random & 0.615 & 0.535 & 0.510 & 0.685 & 0.930 \\
Base & same pred. & 0.615 & 0.535 & 0.620 & 0.965 & 0.965 \\
Base & different pred. & 0.615 & 0.535 & 0.410 & 0.175 & 0.825 \\
\bottomrule
\end{tabular}
\end{table}

\begin{figure}[H]
  \centering
  \includegraphics[width=\linewidth]{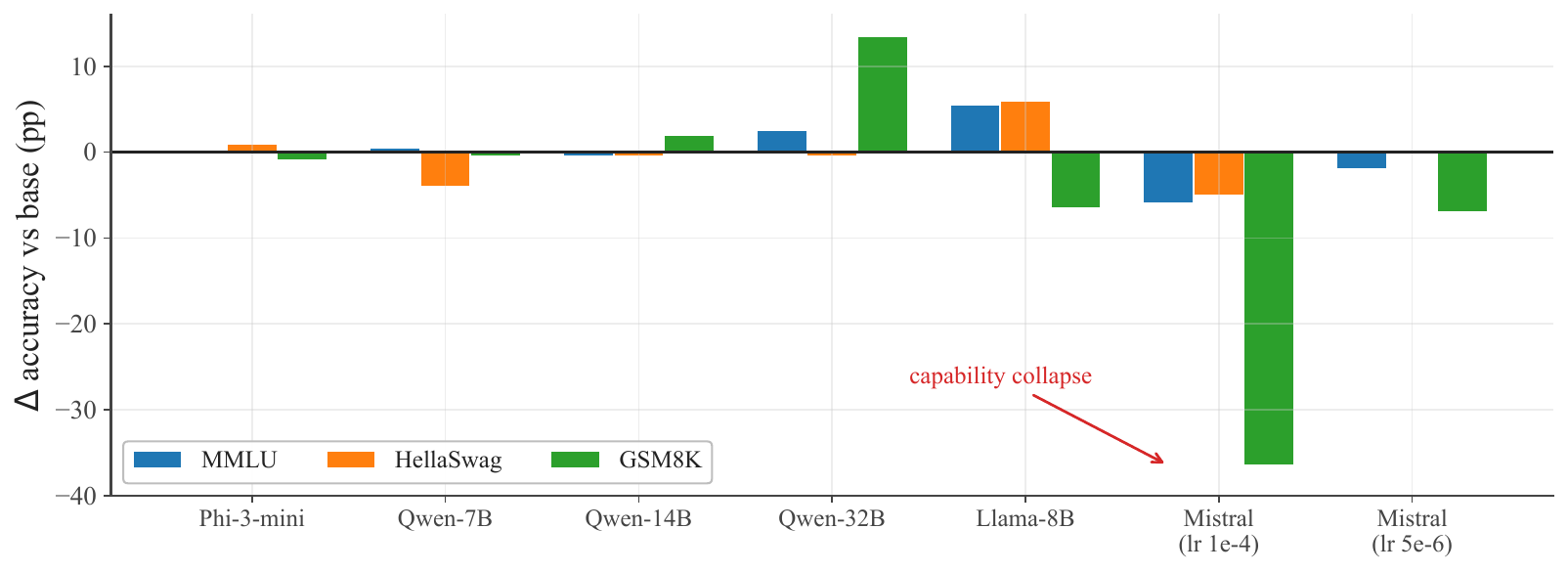}
  \caption{General-capability change after \method{} on MMLU
  \citep{mmlu}, HellaSwag, and GSM8K (\(n = 200\) items each under a fixed
  seed), reported as \(\Delta\) accuracy in percentage points versus base.
  Qwen2.5-32B is evaluated with QLoRA. The Qwen and Phi rows show no broad
  degradation, whereas Mistral-7B under the canonical learning rate
  collapses on GSM8K; the rescued low-rate recipe (step 150) recovers most
  of the loss. The per-benchmark numbers are in \Cref{tab:capability}.}
  \label{fig:capability}
\end{figure}
\begin{table}[t]
\centering
\caption{General-capability evaluation on MMLU \citep{mmlu},
HellaSwag, and GSM8K ($n = 200$ items each under a fixed seed).
Qwen2.5-32B is evaluated with QLoRA. The two Mistral-7B rows compare
the default and rescued training recipes.}
\label{tab:capability}
\small
\setlength{\tabcolsep}{5pt}
\resizebox{\textwidth}{!}{%
\begin{tabular}{l rr r rr r rr r}
\toprule
& \multicolumn{3}{c}{\textbf{MMLU}}
& \multicolumn{3}{c}{\textbf{HellaSwag}}
& \multicolumn{3}{c}{\textbf{GSM8K}} \\
\cmidrule(lr){2-4} \cmidrule(lr){5-7} \cmidrule(lr){8-10}
\textbf{Model} & Base & Vernier & $\Delta$ & Base & Vernier & $\Delta$ & Base & Vernier & $\Delta$ \\
\midrule
Phi-3-mini (3.8B) & $0.655$ & $0.655$ & $\phantom{+}0.0$
                  & $0.795$ & $0.805$ & $+1.0$
                  & $0.815$ & $0.805$ & $-1.0$ \\
Qwen2.5-7B        & $0.685$ & $0.690$ & $+0.5$
                  & $0.830$ & $0.790$ & $-4.0$
                  & $0.705$ & $0.700$ & $-0.5$ \\
Qwen2.5-14B       & $0.780$ & $0.775$ & $-0.5$
                  & $0.855$ & $0.850$ & $-0.5$
                  & $0.820$ & $0.840$ & $+2.0$ \\
Qwen2.5-32B       & $0.825$ & $0.850$ & $+2.5$
                  & $0.900$ & $0.895$ & $-0.5$
                  & $0.740$ & $0.875$ & $+13.5$ \\
Llama-3.1-8B      & $0.565$ & $0.620$ & $+5.5$
                  & $0.645$ & $0.705$ & $+6.0$
                  & $0.830$ & $0.765$ & $-6.5$ \\
Mistral-7B-v0.3, lr=1e-4$^\dagger$ & $0.525$ & $0.465$ & $-6.0$
                  & $0.650$ & $0.600$ & $-5.0$
                  & $0.525$ & $0.160$ & $-36.5$ \\
Mistral-7B-v0.3, lr=5e-6, step 150 & $0.525$ & $0.505$ & $-2.0$
                  & $0.650$ & $0.650$ & $\phantom{+}0.0$
                  & $0.525$ & $0.455$ & $-7.0$ \\
\bottomrule
\end{tabular}
}
\end{table}

\begin{table}[H]
\centering
\caption{Lexical-overlap contamination check for the capability
benchmarks. The 1500 CLadder training prompts are compared with the
200-item MMLU, HellaSwag, and GSM8K samples using long \(n\)-gram
overlap.}
\label{tab:contamination}
\small
\setlength{\tabcolsep}{8pt}
\begin{tabular}{l r rr}
\toprule
\textbf{Benchmark} & \textbf{Items} & \textbf{8-gram overlap} &
\textbf{13-gram overlap} \\
\midrule
MMLU      & 200 & 0.000 & 0.000 \\
HellaSwag & 200 & 0.000 & 0.000 \\
GSM8K     & 200 & 0.000 & 0.000 \\
\bottomrule
\end{tabular}
\end{table}

\begin{table}[H]
\centering
\caption{Hyperparameter sweeps. Top: KL vs MSE consistency form on
Qwen-7B. Middle: LoRA rank on Qwen-7B. Bottom: $\beta$ sweep on
Qwen-14B. Best per column in each block in bold.}
\label{tab:hyperparams}
\small
\setlength{\tabcolsep}{4pt}
\begin{tabular}{l rrr r}
\toprule
\textbf{Configuration} & \textbf{CLadder Gap} & \textbf{CRASS Gap} & \textbf{e-CARE Gap} & \textbf{Avg.\ OOD reduction} \\
\midrule
\multicolumn{5}{l}{\emph{Consistency form on Qwen-7B ($\beta = 1$, $r = 16$):}}\\
Symmetric KL (default)
   & $\mathbf{-2.0}$ & $\mathbf{+25.0}$ & $\mathbf{+17.5}$ & $\mathbf{-25.2\%}$ \\
MSE on answer-subspace logits
   & $-1.5$ & $+24.5$ & $+19.0$ & $-22.8\%$ \\
\midrule
\multicolumn{5}{l}{\emph{LoRA rank on Qwen-7B ($\beta = 1$, symmetric KL, $\alpha_{\mathrm{LoRA}} = 2r$):}}\\
$r = 4$
   & $-2.0$ & $+26.0$ & $+20.0$ & $-18.3\%$ \\
$r = 8$
   & $\mathbf{-4.5}$ & $+23.5$ & $+22.0$ & $-17.6\%$ \\
$r = 16$ (default)
   & $-2.0$ & $+25.0$ & $\mathbf{+17.5}$ & $\mathbf{-25.2\%}$ \\
$r = 32$
   & $-3.0$ & $\mathbf{+21.5}$ & $+22.5$ & $-19.5\%$ \\
\midrule
\multicolumn{5}{l}{\emph{$\beta$ sweep on Qwen-14B ($r = 16$, symmetric KL):}}\\
$\beta = 0.5$
   & $\mathbf{-1.5}$ & $\mathbf{+18.5}\;(\text{-}31\%)$ & $\mathbf{+18.5}\;(\text{-}14\%)$ & $\mathbf{-22\%}$ \\
$\beta = 1.0$ (default)
   & $-1.5$ & $+20.5\;(\text{-}24\%)$ & $+21.0\;(\text{-}2\%)$ & $-13\%$ \\
$\beta = 2.0$
   & $+2.0$ & $+18.5\;(\text{-}31\%)$ & $+22.0\;(\text{+}2\%)$ & $-14\%$ \\
\bottomrule
\end{tabular}
\end{table}

\begin{table}[H]
\centering
\caption{Adversarial-placeholder robustness on held-out CLadder
($n = 200$). Each Vernier adapter is trained only on the canonical
$X_1, X_2, \ldots$ scheme (top block) and evaluated under three
held-out schemes the adapter has never seen.}
\label{tab:adversarial}
\small
\setlength{\tabcolsep}{4pt}
\resizebox{\textwidth}{!}{%
\begin{tabular}{l l rrr rrr rrr}
\toprule
& & \multicolumn{3}{c}{\textbf{Phi-3-mini}}
& \multicolumn{3}{c}{\textbf{Qwen-7B}}
& \multicolumn{3}{c}{\textbf{Qwen-14B}} \\
\cmidrule(lr){3-5} \cmidrule(lr){6-8} \cmidrule(lr){9-11}
\textbf{Scheme} & \textbf{Setting} & $P_0$ & $P_1$ & Gap & $P_0$ & $P_1$ & Gap & $P_0$ & $P_1$ & Gap \\
\midrule
$X_1, X_2, \ldots$
       & Base    & 0.585 & 0.540 & $+4.5$ & 0.625 & 0.545 & $+8.0$ & 0.615 & 0.580 & $+3.5$ \\
       & Vernier & 0.550 & 0.550 & $\phantom{+}0.0$ & 0.870 & 0.890 & $-2.0$ & 0.920 & 0.935 & $-1.5$ \\
\midrule
\texttt{foo, bar, baz, \ldots}
       & Base    & 0.585 & 0.575 & $+1.0$ & 0.625 & 0.555 & $+7.0$ & 0.615 & 0.605 & $+1.0$ \\
       & Vernier & 0.550 & 0.575 & $-2.5$ & 0.870 & 0.855 & $+1.5$ & 0.920 & 0.915 & $+0.5$ \\
\midrule
\texttt{alpha\_a, alpha\_b, \ldots}
       & Base    & 0.585 & 0.565 & $+2.0$ & 0.625 & 0.540 & $+8.5$ & 0.615 & 0.590 & $+2.5$ \\
       & Vernier & 0.550 & 0.570 & $-2.0$ & 0.870 & 0.845 & $+2.5$ & 0.920 & 0.870 & $+5.0$ \\
\midrule
\texttt{VAR\_A, VAR\_B, \ldots}
       & Base    & 0.585 & 0.550 & $+3.5$ & 0.625 & 0.515 & $+11.0$ & 0.615 & 0.530 & $+8.5$ \\
       & Vernier & 0.550 & 0.545 & $+0.5$ & 0.870 & 0.870 & $\phantom{+}0.0$ & 0.920 & 0.775 & $+14.5$ \\
\bottomrule
\end{tabular}
}
\end{table}

\begin{table}[H]
\centering
\caption{Setting A (CLadder-only training) versus Setting B (joint
training on CLadder, CRASS, and e-CARE) on Qwen2.5-7B and
Qwen2.5-14B. Setting B uses $5{,}154$ pooled training items, single
seed. Best CRASS and best e-CARE per model in bold.}
\label{tab:settingB}
\small
\setlength{\tabcolsep}{6pt}
\begin{tabular}{l l rrr r}
\toprule
& & \multicolumn{3}{c}{\textbf{Gap (pp)}} & \\
\cmidrule(lr){3-5}
\textbf{Model} & \textbf{Setting} & \textbf{CLadder} & \textbf{CRASS} & \textbf{e-CARE} & $\Delta_{\text{OOD}}$ \\
\midrule
\multirow{3}{*}{Qwen2.5-7B}
   & Base
   & $+8.0$  & $+34.0$ & $+23.0$ & --- \\
   & Setting A
   & $\mathbf{-2.5_{\pm 0.9}}$ & $+24.8_{\pm 0.3}$ & $+18.8_{\pm 1.5}$
   & $-22\%$ \\
   & Setting B
   & $+0.5$ & $\mathbf{+5.5}$ & $\mathbf{+16.5}$
   & $\mathbf{-56\%}$ \\
\midrule
\multirow{3}{*}{Qwen2.5-14B}
   & Base
   & $+3.5$ & $+27.0$ & $+21.5$ & --- \\
   & Setting A
   & $\mathbf{-1.5}$ & $+20.5$ & $+21.0$ & $-13\%$ \\
   & Setting B
   & $+1.0$ & $\mathbf{+5.5}$ & $\mathbf{+14.0}$
   & $\mathbf{-56\%}$ \\
\bottomrule
\end{tabular}
\end{table}

\begin{table}[H]
\centering
\caption{Paired statistical tests for Qwen2.5-7B. Gaps are $P_0-P_1$
in percentage points. CIs are paired bootstrap 95\% intervals over
items. McNemar tests whether the paired $P_0/P_1$ correctness
asymmetry differs from zero. The final column is the paired reduction
in gap relative to the base row for the same benchmark.}
\label{tab:stats}
\small
\setlength{\tabcolsep}{4pt}
\begin{tabular}{l l r rr r r r r}
\toprule
\textbf{Benchmark} & \textbf{Setting} & $n$ & $P_0$ & $P_1$ &
\textbf{Gap} & \textbf{Gap CI} & \textbf{McNemar $p$} &
\textbf{$\Delta$ gap vs base} \\
\midrule
CLadder & Base & 200 & 62.5 & 54.5 & +8.0 & [+3.0, +13.0] & 0.004 & -- \\
CLadder & Vernier & 200 & 87.0 & 89.0 & -2.0 & [-6.0, +2.0] & 0.454 & +10.0 [+3.5, +17.0] \\
CRASS & Base & 200 & 84.5 & 50.5 & +34.0 & [+26.5, +41.5] & $<10^{-4}$ & -- \\
CRASS & Vernier & 200 & 88.5 & 63.5 & +25.0 & [+18.0, +32.0] & $<10^{-4}$ & +9.0 [+3.0, +15.5] \\
e-CARE & Base & 200 & 81.5 & 58.5 & +23.0 & [+15.5, +30.5] & $<10^{-4}$ & -- \\
e-CARE & Vernier & 200 & 81.0 & 63.5 & +17.5 & [+10.0, +25.0] & $<10^{-4}$ & +5.5 [+0.5, +11.0] \\
\bottomrule
\end{tabular}
\end{table}

\begin{table}[H]
\centering
\caption{Larger CLadder held-out evaluation ($n=1000$). Gaps are
$P_0-P_1$ in percentage points with paired bootstrap 95\% intervals.
The final column reports paired gap reduction relative to the
corresponding base model.}
\label{tab:large-cladder}
\small
\setlength{\tabcolsep}{5pt}
\begin{tabular}{l l r rr r r r}
\toprule
\textbf{Model} & \textbf{Setting} & $n$ & $P_0$ & $P_1$ &
\textbf{Gap} & \textbf{Gap CI} & \textbf{$\Delta$ gap vs base} \\
\midrule
Qwen2.5-7B & Base & 1000 & 55.8 & 53.4 & +2.4 & [+0.0, +4.8] & -- \\
Qwen2.5-7B & Vernier & 1000 & 84.2 & 85.5 & -1.3 & [-3.0, +0.4] & +3.7 [+0.7, +6.7] \\
Qwen2.5-14B & Base & 1000 & 56.8 & 54.0 & +2.8 & [+0.5, +5.2] & -- \\
Qwen2.5-14B & Vernier & 1000 & 86.6 & 87.9 & -1.3 & [-2.7, +0.1] & +4.1 [+1.3, +6.8] \\
\bottomrule
\end{tabular}
\end{table}

\begin{table}[H]
\centering
\caption{Larger out-of-distribution evaluation for Qwen2.5-7B and
Qwen2.5-14B. CRASS uses all 274 prepared items, and e-CARE uses a 1000-item
sample. The adapters are trained only on CLadder. Gap reductions are
relative to the corresponding base row.}
\label{tab:ood-bign}
\small
\setlength{\tabcolsep}{5pt}
\begin{tabular}{l l c cc c c}
\toprule
\textbf{Model} & \textbf{Benchmark / Setting} & $\bm{n}$ & $\bm{P_0}$ & $\bm{P_1}$ &
\textbf{Gap} & \textbf{$\Delta$ Gap} \\
\midrule
\multirow{4}{*}{Qwen2.5-7B}
  & CRASS Base    & 274 & 0.839 & 0.489 & $+35.0$ & -- \\
  & CRASS Vernier & 274 & 0.876 & 0.639 & $\mathbf{+23.7}$ & $\mathbf{-32\%}$ \\
  & e-CARE Base    & 1000 & 0.802 & 0.595 & $+20.7$ & -- \\
  & e-CARE Vernier & 1000 & 0.808 & 0.621 & $\mathbf{+18.7}$ & $\mathbf{-10\%}$ \\
\midrule
\multirow{4}{*}{Qwen2.5-14B}
  & CRASS Base    & 274 & 0.894 & 0.620 & $+27.4$ & -- \\
  & CRASS Vernier & 274 & 0.923 & 0.755 & $\mathbf{+16.8}$ & $\mathbf{-39\%}$ \\
  & e-CARE Base    & 1000 & 0.824 & 0.592 & $+23.2$ & -- \\
  & e-CARE Vernier & 1000 & 0.830 & 0.606 & $+22.4$ & $-3\%$ \\
\bottomrule
\end{tabular}
\end{table}

\begin{table}[H]
\centering
\caption{Base-model CLadder lexical gaps on a larger held-out slice
(\(n=1000\)). Gaps are \(P_0-P_1\) in percentage points with paired
bootstrap 95\% intervals, and \(p\) is the exact two-sided McNemar test on
paired correctness.}
\label{tab:base-gap-significance}
\small
\setlength{\tabcolsep}{6pt}
\begin{tabular}{l cc c c c}
\toprule
\textbf{Model} & \(\bm{P_0}\) & \(\bm{P_1}\) & \textbf{Gap} & \textbf{95\% CI} & \textbf{McNemar \(p\)} \\
\midrule
Qwen2.5-7B    & 0.553 & 0.534 & \(+1.9\) & \([-0.6,+4.4]\) & 0.151 \\
Qwen2.5-14B   & 0.564 & 0.549 & \(+1.5\) & \([-0.9,+3.9]\) & 0.255 \\
Qwen2.5-32B   & 0.568 & 0.558 & \(+1.0\) & \([-1.3,+3.3]\) & 0.440 \\
Llama-3.1-8B  & 0.530 & 0.508 & \(+2.2\) & \([-1.0,+5.3]\) & 0.194 \\
\bottomrule
\end{tabular}
\end{table}

\begin{table}[H]
\centering
\caption{Random stratified CLadder held-out splits on Qwen2.5-7B. Each run
trains a fresh adapter on 1500 training items and evaluates on a disjoint
200-item held-out set stratified by rung. Values are mean \(\pm\) standard
deviation across split seeds \(\{7,42,123\}\).}
\label{tab:stratified-splits}
\small
\setlength{\tabcolsep}{7pt}
\begin{tabular}{l ccc}
\toprule
\textbf{Aggregate} & \textbf{Base Gap} & \textbf{Vernier Gap} & \textbf{\(\Delta\) Gap} \\
\midrule
Mean \(\pm\) std & +6.2 $\pm$ 5.2 & \textbf{+0.2 $\pm$ 1.4} & \textbf{+6.0 $\pm$ 5.0} \\
\bottomrule
\end{tabular}
\end{table}

\begin{table}[H]
\centering
\caption{Capacity and mechanism sweep on held-out CLadder ($n = 200$).
The gap is $P_0 - P_1$. ``Probe $\Delta$'' is the change in final-layer
$P_1$ probe-transfer accuracy, and ``lens ratio'' is the base-to-\method{}
logit-lens disagreement ratio. Small Qwen rows ($\dagger$) are
three-seed means $\pm$ std, and other rows are single-seed.}
\label{tab:capacity}
\small
\setlength{\tabcolsep}{5pt}
\begin{tabular}{l r cc c c l}
\toprule
\textbf{Model} & \textbf{Params} & \textbf{Base Gap} & \textbf{Vernier Gap} & \textbf{Probe $\Delta$} & \textbf{Lens Ratio} & \textbf{Verdict} \\
\midrule
Qwen2.5-1.5B$^\dagger$ & 1.5B & -9.0 & +3.5 & -0.09 & $1.1\times$ & fail \\
Qwen2.5-3B$^\dagger$   & 3B   & +0.5 & +0.5 & \textbf{+0.29} & $2.1\times$ & \textbf{success} \\
Phi-3-mini   & 3.8B & +4.5 & +0.0 & +0.04 & $1.0\times$ & fail \\
Gemma-2-2B   & 2B   & +9.5 & -2.0 & +0.07 & $0.4\times$ & mixed \\
\midrule
Qwen2.5-7B   & 7B   & +8.0 & -2.5 & +0.22 & $5.5\times$ & success \\
Qwen2.5-14B  & 14B  & +3.5 & +0.5 & +0.09 & $5.0\times$ & success \\
Qwen2.5-32B  & 32B  & -2.0 & +0.5 & +0.12 & $3.1\times$ & success \\
\bottomrule
\end{tabular}
\end{table}

\begin{table}[H]
\centering
\caption{e-CARE final-layer decision-token alignment. Cosine is
\(\cos(h_L(P_0),h_L(P_1))\) at the last non-pad token. The remaining
e-CARE behavioural gap is therefore not explained by a failure to move
the two final-layer representations closer together.}
\label{tab:ecare-mech}
\small
\setlength{\tabcolsep}{7pt}
\begin{tabular}{l ccc}
\toprule
\textbf{Model} & \textbf{Base Cosine} & \textbf{Vernier Cosine} &
\textbf{\(\Delta\)} \\
\midrule
Qwen2.5-7B  & 0.904 & 0.923 & +0.019 \\
Qwen2.5-14B & 0.917 & 0.936 & +0.020 \\
Qwen2.5-32B & 0.902 & 0.924 & +0.021 \\
\bottomrule
\end{tabular}
\end{table}

\begin{table}[H]
\centering
\caption{Preliminary non-causal BBH checks with object names
renamed to typed placeholders. Logical\_deduction uses $n = 100$
held-out items; colored\_objects uses a smaller $n = 20$ diagnostic
slice. Augmentation closes or reduces the gap with both views rising,
and activation patching shows the decision-token representation is a
control point in the tested cells.}
\label{tab:logdeduct}
\small
\setlength{\tabcolsep}{6pt}
\begin{tabular}{l l ccc ccc}
\toprule
& & \multicolumn{3}{c}{\textbf{Base}}
& \multicolumn{3}{c}{\textbf{After Augmentation}} \\
\cmidrule(lr){3-5} \cmidrule(lr){6-8}
\textbf{Task} & \textbf{Model} & $\bm{P_0}$ & $\bm{P_1}$ & \textbf{Gap} & $\bm{P_0}$ & $\bm{P_1}$ & \textbf{Gap} \\
\midrule
\multirow{2}{*}{Logical deduction}
  & Qwen-7B  & 0.740 & 0.650 & +9.0 & 0.950 & 0.950 & \textbf{+0.0} \\
  & Qwen-14B & 0.900 & 0.890 & +1.0 & 0.990 & 0.980 & +1.0 \\
\midrule
\multirow{2}{*}{Colored objects}
  & Qwen-7B  & 0.500 & 0.300 & +20.0 & 0.800 & 0.850 & \textbf{-5.0} \\
  & Qwen-14B & 0.750 & 0.300 & +45.0 & 0.850 & 0.650 & +20.0 \\
\midrule
\multicolumn{8}{l}{\textbf{Activation patching} ($P_0$ decision-token state copied into $P_1$)} \\
\multicolumn{8}{l}{\quad logical deduction, Qwen-7B, depth 0.8: $P_1$ $0.63\!\to\!\mathbf{0.77}$, flip-to-$P_0$ $\mathbf{0.89}$} \\
\multicolumn{8}{l}{\quad colored objects, Qwen-14B, depth 0.8: $P_1$ $0.25\!\to\!\mathbf{0.75}$, flip-to-$P_0$ $\mathbf{1.00}$} \\
\bottomrule
\end{tabular}
\end{table}

\begin{figure}[H]
  \centering
  \includegraphics[width=0.82\linewidth]{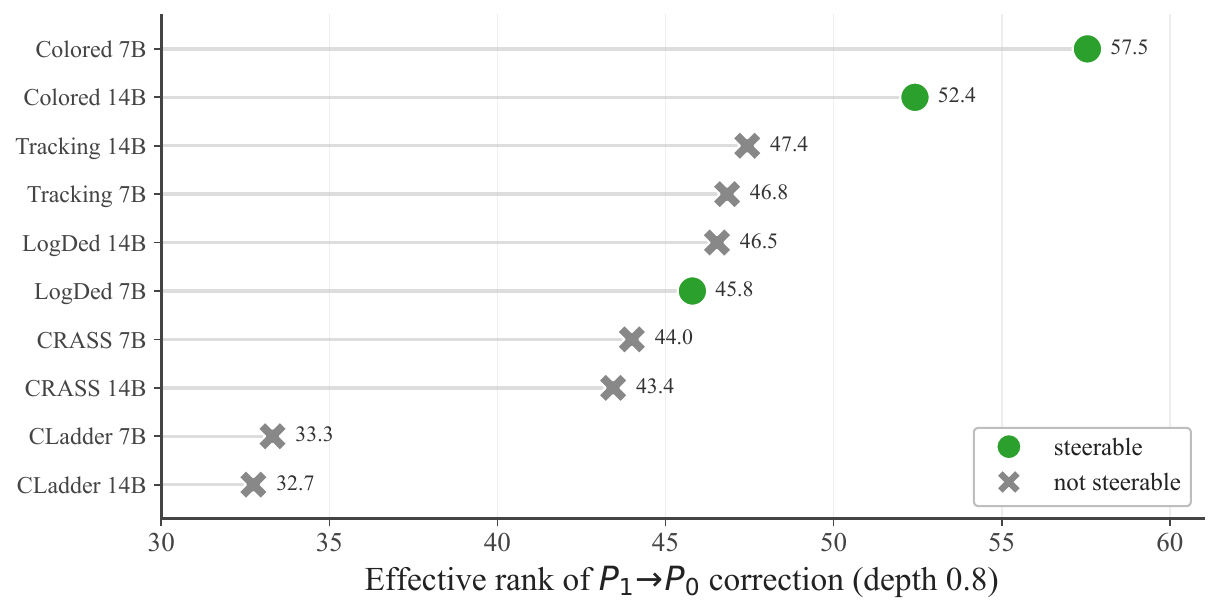}
  \caption{Effective rank does not explain task-dependent steerability.
  Effective rank is the participation ratio of the \(\Pone\!\to\!\Pzero\)
  linear-correction spectrum at depth 0.8, and a task is marked steerable
  if a swept linear or MLP map closes its gap. Steerable (green) and
  non-steerable (grey) tasks interleave along the rank axis: CLadder has a
  lower rank than the steerable non-causal tasks yet stays unsteerable,
  while colored objects has the highest ranks and is steerable on both
  Qwen scales. Base gaps are measured on the steering map-fitting slice
  and can differ by a few points from the held-out gaps in
  \Cref{tab:logdeduct} (logical\_deduction uses \(n = 100\), colored
  objects uses \(n = 20\)). The underlying values are in \Cref{tab:effrank}.}
  \label{fig:effrank}
\end{figure}
\begin{table}[H]
\centering
\caption{Effective rank does not explain task-dependent steerability.
Effective rank is the participation ratio of the \(P_1\!\to P_0\)
linear-correction spectrum at depth 0.8. Best gap is the smallest
residual gap reached by any swept linear or MLP map. Base gaps here are
measured on the steering map-fitting slice and can differ by a few
points from the held-out gaps in \Cref{tab:logdeduct} (logical\_deduction
uses $n = 100$, while colored objects uses $n = 20$; e.g. colored
objects on Qwen-14B is $+50.0$ here versus $+45.0$ held-out).}
\label{tab:effrank}
\small
\setlength{\tabcolsep}{6pt}
\begin{tabular}{l l c cc c}
\toprule
\textbf{Task} & \textbf{Model} & \textbf{Eff. Rank}
& \textbf{Base Gap} & \textbf{Best Gap} & \textbf{Steerable} \\
\midrule
CLadder & qwen25-14b & 32.74 & +4.0 & +4.5 & $\times$ \\
CLadder & qwen25-7b & 33.31 & +5.5 & +2.5 & $\times$ \\
CRASS & qwen25-14b & 43.44 & +29.0 & +29.8 & $\times$ \\
CRASS & qwen25-7b & 44.00 & +41.1 & +29.0 & $\times$ \\
logical deduction & qwen25-7b & 45.80 & +9.0 & +1.0 & \checkmark \\
logical deduction & qwen25-14b & 46.53 & +0.0 & +0.0 & $\times$ \\
tracking objects & qwen25-7b & 46.83 & -9.0 & +1.0 & $\times$ \\
tracking objects & qwen25-14b & 47.43 & -1.0 & -1.0 & $\times$ \\
colored objects & qwen25-14b & 52.42 & +50.0 & +20.0 & \checkmark \\
colored objects & qwen25-7b & 57.55 & +20.0 & +5.0 & \checkmark \\
\bottomrule
\end{tabular}
\end{table}

\begin{figure}[H]
  \centering
  \includegraphics[width=0.72\linewidth]{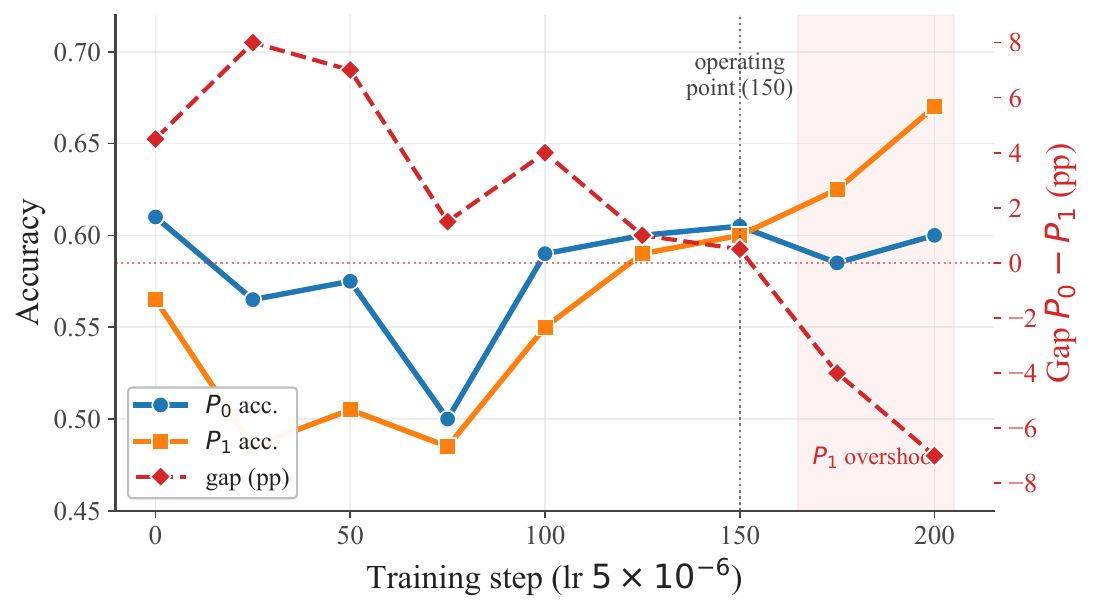}
  \caption{Mistral-7B-Instruct-v0.3 checkpoint trajectory under the
  rescued recipe (lr \(5\!\times\!10^{-6}\)) on the in-distribution
  CLadder held-out slice. \(\Pzero\) and \(\Pone\) accuracy (left axis)
  and the signed gap \(\Pzero - \Pone\) (right axis) are plotted against
  the training step. Step 150 is the operating point reported in
  \Cref{tab:main,tab:ood,fig:capability}; later checkpoints overshoot into
  a negative gap as \(\Pone\) overtakes \(\Pzero\). The per-step numbers are
  in \Cref{tab:mistral-sweep}.}
  \label{fig:mistral-sweep}
\end{figure}
\begin{table}[H]
\centering
\caption{Mistral-7B-Instruct-v0.3 checkpoint trajectory under the
rescued recipe (lr=$5\!\times\!10^{-6}$) on the in-distribution
CLadder held-out slice. Step 150 is the operating point reported in
\Cref{tab:main,tab:ood,tab:capability}.}
\label{tab:mistral-sweep}
\small
\setlength{\tabcolsep}{6pt}
\begin{tabular}{l ccc l}
\toprule
\textbf{Step} & $\bm{P_0}$ & $\bm{P_1}$ & \textbf{CLadder gap} & \textbf{Comment} \\
\midrule
Base   & 0.610 & 0.565 & +4.5 & --- \\
\midrule
25  & 0.565 & 0.485 & +8.0 & no learning yet \\
50  & 0.575 & 0.505 & +7.0 & \\
75  & 0.500 & 0.485 & +1.5 & \\
100 & 0.590 & 0.550 & +4.0 & partial closure \\
125 & 0.600 & 0.590 & +1.0 & \\
150 & 0.605 & 0.600 & \textbf{+0.5} & sweet spot, reported in \Cref{tab:main,tab:ood,tab:capability} \\
175 & 0.585 & 0.625 & -4.0 & $P_1$ overshoot \\
200 & 0.600 & 0.670 & -7.0 & $P_1$ overshoot \\
\bottomrule
\end{tabular}
\end{table}


\section{Pre-specification, recipe audit, and falsification summary}
\label{sec:audit}

To make the selection process auditable, we list the default recipe and
every per-model deviation, together with the trigger and the metric used
to choose it. The default recipe uses LoRA
rank 16 on the attention projections, \(\alpha=\beta=1\), learning rate
\(1\times10^{-4}\), and 500 steps. Two models deviate. Phi-3-mini uses
learning rate \(5\times10^{-6}\) with early stopping at step 70, because
the canonical learning rate drives the adapter to a constant-output
minimum by step 50; the trigger is the degenerate in-distribution
collapse, observed before any out-of-distribution evaluation.
Mistral-7B uses the same low learning rate with the operating point at
step 150, selected as the earliest checkpoint whose in-distribution gap
is within 1 pp of zero while avoiding the capability collapse of the
canonical run (\Cref{fig:mistral-sweep}); out-of-distribution metrics
are not used for this selection. The 14B and 32B models use QLoRA for
memory reasons only, with no change to the loss or selection rule. Seed
choices are fixed in advance to \(\{7,42,123\}\) for the three-seed
models and to 42 for the single-seed models; no seed is dropped.

The argument is organised so that each diagnostic can fail
independently. \Cref{tab:falsifiers} summarises the prediction each
account makes, the observation that would falsify the
representational-misalignment reading, and where the outcome is
reported. The misalignment account survives on the working-regime
models and is falsified, by these same criteria, on Phi-3 and Gemma,
which is why the later claims are scoped by capacity and base-model
identity rather than asserted universally.

\begin{table}[H]
\centering
\caption{Falsification summary. Each row is an independent test of the
representational-misalignment account against simple information
erasure or a generic-consistency explanation.}
\label{tab:falsifiers}
\small
\setlength{\tabcolsep}{4pt}
\resizebox{\textwidth}{!}{%
\begin{tabular}{l l l}
\toprule
\textbf{Prediction (misalignment)} & \textbf{Falsifier} & \textbf{Where} \\
\midrule
Gap closes with \(\Pone\) rising, not \(\Pzero\) collapsing
  & Closure driven by \(\Pzero\) regression
  & \Cref{tab:main} (Phi-3 falsifies) \\
Story content stays recoverable from \(\Pone\)
  & Probe transfer falls toward chance
  & \Cref{tab:mech} (probe rises) \\
Decision-token beliefs of the two views converge
  & Logit-lens disagreement does not move
  & \Cref{tab:mech} (Phi-3 falsifies) \\
\(T\), not generic consistency, drives closure
  & R-Drop (no \(T\)) reproduces the effect
  & \Cref{tab:ablation} (R-Drop widens gap) \\
The decision-token state carries answer identity
  & Matched patch is no better than random donor
  & \Cref{fig:patch,fig:patch-controls} \\
The gap is item-conditional, not a fixed direction
  & A once-fit map removes it at inference
  & \Cref{tab:align,tab:steer-exhaustive} (map fails) \\
\bottomrule
\end{tabular}
}
\end{table}


\section{Phi-3 with MLP-targeted LoRA}
\label{sec:phi3-mlp}

The default \method{} configuration places LoRA on the attention projections. On Phi-3 this drives the in-distribution gap to zero through \(\Pzero\) regression and leaves the hidden representation unchanged (\Cref{sec:mech}). To test whether the failure is intrinsic to Phi-3 or specific to the attention-only LoRA target, we re-train Phi-3 with the LoRA target set switched to the MLP projections, keeping the loss, optimiser, learning rate (\(5 \times 10^{-6}\)), batch size, rank, and number of steps identical, and sweep the checkpoint at step \(\in \{25, 50, 75, 100\}\) on the in-distribution held-out slice. The full trajectory together with out-of-distribution transfer is reported in \Cref{tab:phi3mlp}.

\begin{table}[H]
\centering
\caption{Phi-3-mini with the LoRA target switched from attention to
MLP projections (\texttt{gate\_up\_proj}, \texttt{down\_proj}),
keeping all other hyperparameters identical. The default
attention-LoRA row is included for reference.}
\label{tab:phi3mlp}
\small
\setlength{\tabcolsep}{5pt}
\begin{tabular}{l l ccc r}
\toprule
\textbf{LoRA target} & \textbf{Step} & $\bm{P_0}$ & $\bm{P_1}$ & \textbf{CLadder gap} & \textbf{OOD reduction} \\
\midrule
Base (no adapter) & ---
   & 0.585 & 0.540 & +4.5 & --- \\
\midrule
\multicolumn{6}{l}{\emph{Attention projections (\texttt{q,k,v,o\_proj}), default \method{}:}}\\
\texttt{attn}  & 500
   & 0.550 & 0.550 & 0.0 & CRASS -3\%, e-CARE -10\% \\
\midrule
\multicolumn{6}{l}{\emph{MLP projections (\texttt{gate\_up\_proj}, \texttt{down\_proj}):}}\\
\texttt{mlp}  & 25  & 0.580 & 0.545 & +3.5 & --- \\
\texttt{mlp}  & 50  & 0.570 & 0.535 & +3.5 & CRASS -3\%, e-CARE -7\% \\
\texttt{mlp}  & 75  & 0.510 & 0.490 & +2.0 & CRASS +2\%, e-CARE -5\% \\
\texttt{mlp}  & 100 & 0.490 & 0.490 & 0.0 & CRASS -5\%, e-CARE -2\% \\
\bottomrule
\end{tabular}
\end{table}

The MLP-LoRA variant reproduces the attention-LoRA failure with no qualitative difference. The in-distribution gap does close (to 2.0 pp at step 75 and to 0.0 pp at step 100), but \(\Pzero\) regresses monotonically from 0.585 at base to 0.490 at step 100, the chance rate on the binary-answer subset. The OOD gaps on CRASS and e-CARE stay within 3 pp of the base model at every checkpoint. Switching the LoRA target therefore does not change the conclusion of \Cref{sec:scale}. In our runs, the 3.8B adapter does not find an answer-subspace invariance that also preserves task accuracy, and the optimiser falls back to the same confidence-suppression shortcut regardless of which projection family it acts on.


\section{Per-rung CLadder breakdown}
\label{sec:perrung}

The held-out CLadder results broken down by Pearl-ladder rung are reported in \Cref{tab:rung}.

\begin{table}[H]
\centering
\caption{Per-rung CLadder breakdown ($n = 200$ held-out), grouped by model so the three rungs can be compared within each model. Rung $n$ is the count of held-out items at that rung. ``flip'' marks a base gap that reverses sign after \method{} (relative reduction undefined). $\Delta_{\text{gap}}$ is the relative gap reduction versus base. Within each model, \textbf{bold} marks the best and \underline{underline} the second best across the three rungs per column ($P_0$/$P_1$ higher is better; Gap closest to $0$ is better). $\Delta_{\text{gap}}$ is shown but not ranked, because the per-rung base gaps differ and flips and overcorrection make it non-monotonic.}
\label{tab:rung}
\small
\setlength{\tabcolsep}{4pt}
\begin{tabular}{c l ccc ccc c}
\toprule
& & \multicolumn{3}{c}{\textbf{Base}} & \multicolumn{3}{c}{\textbf{Vernier}} & \\
\cmidrule(lr){3-5} \cmidrule(lr){6-8}
\textbf{Model} & \textbf{Rung} ($n$) & $\bm{P_0}$ & $\bm{P_1}$ & \textbf{Gap} & $\bm{P_0}$ & $\bm{P_1}$ & \textbf{Gap} & \textbf{Rel. gap} \\
\midrule
\multirow{3}{*}{\rotatebox{0}{Phi-3-mini}}
  & Observational (75)   & 0.547 & \underline{0.520} & \textbf{+2.7} & 0.520 & \underline{0.547} & \underline{-2.7} & flip \\
  & Interventional (57)  & \textbf{0.667} & \textbf{0.614} & \underline{+5.3} & \textbf{0.596} & \textbf{0.614} & \textbf{-1.8} & flip \\
  & Counterfactual (68)  & \underline{0.559} & 0.500 & +5.9 & \underline{0.544} & 0.500 & +4.4 & -25\% \\
\midrule
\multirow{3}{*}{\rotatebox{0}{Qwen2.5-7B}}
  & Observational (75)   & 0.587 & \underline{0.547} & \textbf{+4.0} & \underline{0.880} & \underline{0.853} & \textbf{+2.7} & -32\% \\
  & Interventional (57)  & \textbf{0.684} & \textbf{0.579} & +10.5 & \textbf{0.947} & \textbf{1.000} & -5.3 & -150\% \\
  & Counterfactual (68)  & \underline{0.618} & 0.515 & \underline{+10.3} & 0.794 & 0.838 & \underline{-4.4} & -143\% \\
\midrule
\multirow{3}{*}{\rotatebox{0}{Qwen2.5-14B}}
  & Observational (75)   & \underline{0.587} & 0.573 & \textbf{+1.3} & 0.893 & 0.907 & \textbf{-1.3} & flip \\
  & Interventional (57)  & \textbf{0.632} & \textbf{0.596} & \underline{+3.5} & \textbf{0.947} & \textbf{1.000} & -5.3 & -251\% \\
  & Counterfactual (68)  & \textbf{0.632} & \underline{0.574} & +5.9 & \underline{0.926} & \underline{0.912} & \underline{+1.5} & -75\% \\
\bottomrule
\end{tabular}
\end{table}

On both Qwen models the largest baseline gap is on the interventional and counterfactual rungs and the smallest is on the observational rung. After \method{} the interventional \(\Pone\) accuracy reaches 1.000 on both Qwen models, the largest single-cell accuracy improvement reported in the paper, consistent with the hypothesis that lexical anchors substitute most heavily for causal-structure inference on the two causal rungs. Phi-3 shows the opposite pattern. The interventional rung gap closes purely through \(\Pzero\) regression, with \(\Pzero\) falling from 0.667 to 0.596 while \(\Pone\) stays at 0.614. This is the per-rung signature of the mechanism null reported above.

\end{document}